\documentclass{article}

    \PassOptionsToPackage{numbers, compress}{natbib}

    \usepackage[preprint]{neurips_2022}

\usepackage[utf8]{inputenc} 
\usepackage[T1]{fontenc}    
\usepackage{hyperref}       
\usepackage{url}            
\usepackage{booktabs}       
\usepackage{amsfonts}       
\usepackage{nicefrac}       
\usepackage{microtype}      
\usepackage{xcolor}         
\usepackage{overpic}
\usepackage{enumitem} 
\usepackage{overpic} 
\usepackage{color}
\usepackage{amsmath}
\usepackage{bbm}
\usepackage{amssymb}

\definecolor{turquoise}{cmyk}{0.65,0,0.1,0.3}
\definecolor{purple}{rgb}{0.65,0,0.65}
\definecolor{dark_green}{rgb}{0, 0.5, 0}
\definecolor{orange}{rgb}{0.8, 0.6, 0.2}
\definecolor{red}{rgb}{0.8, 0.2, 0.2}
\definecolor{darkred}{rgb}{0.6, 0.1, 0.05}
\definecolor{blueish}{rgb}{0.0, 0.3, .6}
\definecolor{light_gray}{rgb}{0.7, 0.7, .7}
\definecolor{pink}{rgb}{1, 0, 1}
\definecolor{greyblue}{rgb}{0.25, 0.25, 1}

\newcommand{\loss}[1]{\mathcal{L}_\text{#1}}

\newcommand{\real}{\mathbb{R}}
\newcommand{\PC}{\mathbf{P}}
\newcommand{\R}{\mathbf{R}}
\newcommand{\T}{\mathbf{T}}
\newcommand{\J}{\mathcal{J}}
\newcommand{\M}{\mathcal{M}}
\newcommand{\x}{\mathbf{x}}
\newcommand{\y}{\mathbf{y}}
\newcommand{\p}{\mathbf{p}}

\usepackage{blindtext}

\renewcommand{\paragraph}[1]{\vspace{1em}\noindent\textbf{#1}.}

\title{Tracking and Reconstructing Hand Object Interactions from Point Cloud Sequences in the Wild}

\author{%
Jiayi Chen*\textsuperscript{1,2} \quad
Mi Yan*\textsuperscript{1} \quad 
Jiazhao Zhang\textsuperscript{1} \quad 
Yinzhen Xu\textsuperscript{1,2} \quad 
Xiaolong Li\textsuperscript{3} \\ 
\textbf{Yijia Weng\textsuperscript{4} \quad 
Li Yi \textsuperscript{5} \quad 
Shuran Song \textsuperscript{6} \quad
He Wang\textsuperscript{1$^\dagger$}} \\
\textsuperscript{1}CFCS, Peking University \quad 
\textsuperscript{2}Beijing Institute for General AI \quad
\textsuperscript{3}Virginia Tech \\
\textsuperscript{4}Stanford University \quad
\textsuperscript{5}Tsinghua University \quad 
\textsuperscript{6}Columbia University\quad
}

\begin{document}

\setcitestyle{square}
\maketitle

\begin{abstract}
\vspace{-3mm}
In this work, we tackle the challenging task of jointly tracking hand object pose and reconstructing their shapes from depth point cloud sequences in the wild, given the initial poses at frame 0.
We for the first time propose a point cloud based hand joint tracking network, HandTrackNet, to estimate the inter-frame hand joint motion. Our HandTrackNet proposes a novel hand pose canonicalization module to ease the tracking task, yielding accurate and robust hand joint tracking. Our pipeline then reconstructs the full hand via converting the predicted hand joints into a template-based parametric hand model MANO.
For object tracking, we devise a simple yet effective module that estimates the object SDF from the first frame and performs optimization-based tracking.
Finally, a joint optimization step is adopted to perform joint hand and object reasoning, which alleviates the occlusion-induced ambiguity and further refines the hand pose.
During training, the whole pipeline only sees purely synthetic data, which are synthesized with sufficient variations and by depth simulation for the ease of generalization.
The whole pipeline is pertinent to the generalization gaps and thus directly transferable to real in-the-wild data.
We evaluate our method on two real hand object interaction datasets, \textit{e.g.} HO3D and DexYCB, without any finetuning. 
Our experiments demonstrate that the proposed method significantly outperforms the previous state-of-the-art depth-based hand and object pose estimation and tracking methods, running at a frame rate of 9 FPS. 
\end{abstract}
\vspace{-3mm}

\let\thefootnote\relax\footnote{$^*$: Equal contribution, $^\dagger$: Corresponding author, \texttt{hewang@pku.edu.cn}}

\begin{figure}[htp]
\centering
\includegraphics[width=1\columnwidth]{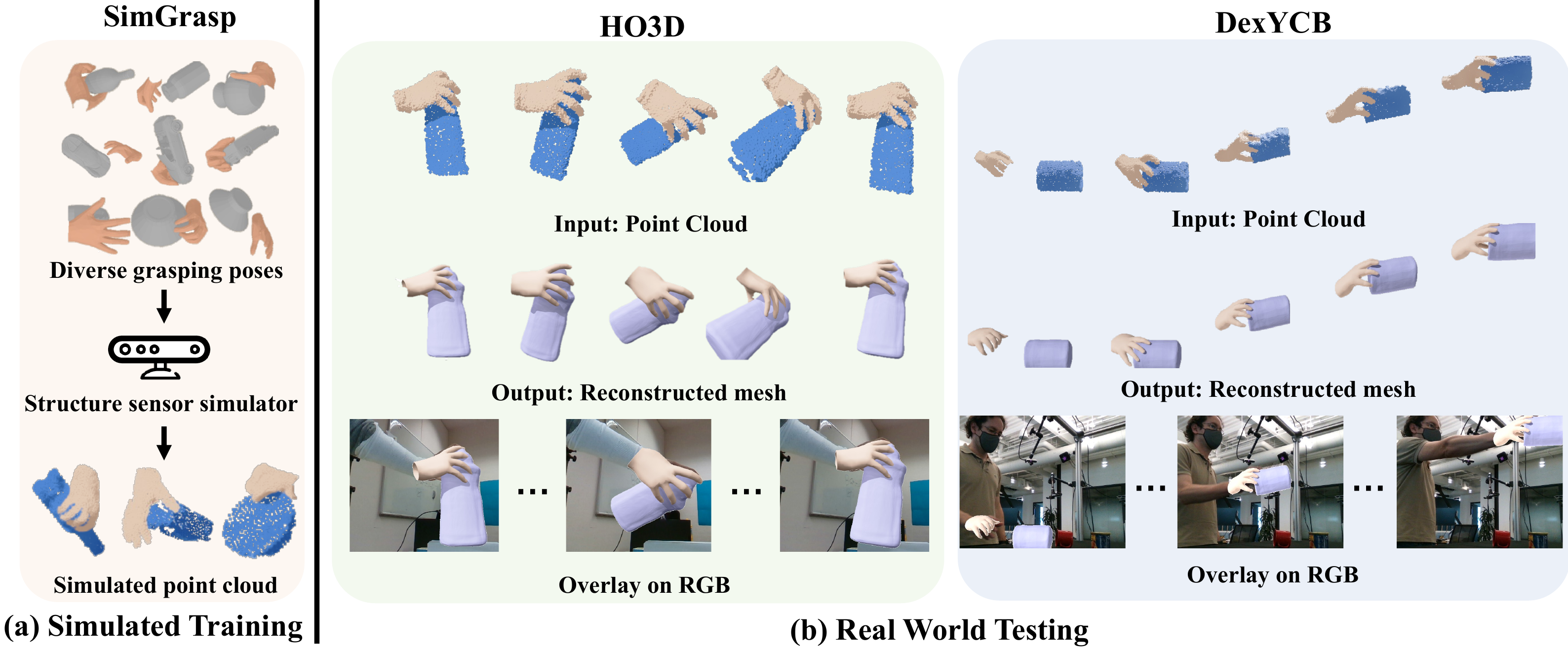}

\caption{Left: We generate a large-scale hand-object interaction dataset, named SimGrasp, using simulated structure light based depth sensor. Right: Trained only on SimGrasp, our methods can be directly transferred to the challenging real world datasets, \textit{i.e.} HO3D\cite{hampali2020ho3dv2} and DexYCB\cite{chao2021dexycb}, to track and reconstruct  hand object interaction.}

\end{figure}
\vspace{-4mm}
\section{Introduction}
\vspace{-4mm}
Hand object interactions (HOI) are ubiquitous in our daily life and form a major way for our humans to interact with complex real-world scenes. 
As major approaches to perceive human object interactions, 
pose tracking and reconstructing human object interaction in 3D are crucial research topics and can enable a broad range of applications, including human-computer interaction \cite{sridhar2015investigating}, human-robot interaction \cite{yang2020human}, augmented reality \cite{hurst2013gesture,piumsomboon2013user}, and learning from human demonstrations \cite{garciahernando2020physics,handa2020dexpilot,qin2021dexmv}.  

Driven by the great power of deep learning, recent years have witnessed much progress in developing learning-based methods for perceiving hand and object under a single frame setting, \textit{e.g.} 3D hand pose estimation from single RGB images ~\cite{oberweger2015hands,madadi2017end,oberweger2017deepprior++,guo2017towards,guo2017region,zhang2019pixel} and point clouds ~\cite{ge2018hand,li2019point,cheng2021handfoldingnet} , instance-level and category-level 6D object pose estimation ~\cite{he2020pvn3d,wang2019normalized}, and joint hand object pose estimation~\cite{doosti2020hopenet, tekin2019h+o} and reconstruction~\cite{hasson2019obman, zhang2021single, hasson21homan} for hand object interaction.

Despite many applications require temporally coherent estimations and thus prefer tracking rather than single-frame estimation,
hand object pose tracking is still an under-explored area for learning-based methods. 
One obvious reason is the lack of large-scale annotated video data, given that only very few small-scale HOI video datasets exist, e.g. DexYCB\cite{chao2021dexycb} and HO3D\cite{hampali2020ho3dv2}, which only cover very limited variations. 
Especially, a useful learning-based hand object tracking method needs to generalize well to novel hand and object instances in novel motion trajectories, which further aggravates the data issue. Apparently, curating a large-scale fully-annotated 3D hand object interaction video dataset would be a valid step. However, the cost of doing so would be formidable, especially for 3D annotations.

In this work, we propose a novel framework to tackle this very challenging task:  \textit{jointly track pose and reconstruct the hand and object from depth point cloud sequences in the wild without seeing any real data during training}. Our setting is as follows: given a depth point cloud sequence with segmented hand and object along with initial hand pose and object pose, our system needs to recover hand object shapes and track their poses in an online manner. We choose point cloud over images since point cloud contains metric 3D geometry of the hand and object, which enables us to obtain full 3D hand and object poses and shapes, and arguably have fewer ambiguities. 

To this ambitious objective, there are several major challenges. The biggest challenges come from generalization: 1) the tracking network needs to generalize well across the huge spatial and temporal variations in hand object interactions; and 2) the Sim2Real gap, due to no real training data. Also, during hand object interaction the heavy inter- and self-occlusions may result in many ambiguities and thus high ill-posedness, leading to further learning challenges.

To mitigate the data issue, we propose a simulation pipeline to synthesize diverse hand object interactions and carry free annotations of their shapes and poses. To minimize the Sim2Real gap, we leverage the structure light based depth sensor simulator proposed by DDS \cite{planche2021physics} to generate simulated depths that carry realistic sensor noises. 

Using purely the simulated data for training, we for the first time propose a point cloud based hand pose tracking network, namely HandTrackNet, to track the inter-frame hand joint motion. HandTrackNet is built upon PointNet++~\cite{qi2017pointnet++} and can estimate the joint positions based on the predictions from the last frame. Based on the temporal continuity, it extracts the hand global pose from the latest prediction and uses it to canonicalize the hand point cloud, which significantly eases the regression task. During training, it learns to track random inter-frame joint motions and is thus free from overfitting to any temporal trajectories. 

To track a novel object, we use DeepSDF\cite{park2019deepsdf}, which leverages category shape prior, to reconstruct the full geometry of the unseen object at the very first frame, and then simply perform a depth-to-SDF conformation based optimization to track pose. Though experiments we show that this simple method can already work well and allow generalization to novel object category with similar shape, \textit{e.g.} trained on bottle and tested on box and can. Compared to previous works \cite{liu2021semi,grady2021contactopt} about HOI, most of which can only track known objects, we have already token a great step towards generalization.

Finally, to alleviate the ambiguities and complexity in hand object interaction, we leverage optimization-based approaches to reason the spatial relationship between the hand and object. We turn the tracked hand joint positions into a MANO~\cite{romero2017embodied} hand representation and construct several energy functions based on HOI priors, to yield more physically plausible hand object poses.
Our extensive experiments demonstrate the effectiveness of our method on never seen real-world hand object interaction datasets, HO3D\cite{hampali2020ho3dv2} and DexYCB\cite{chao2021dexycb}. Trained only on our simulated data, our method outperforms previous approaches on both hand and object pose tracking, and shows good tracking robustness and great generalizability.

In summary, our contribution lies in three aspects: (1) We propose an online hand object pose tracking and reconstruction system taking inputs point cloud sequences under the challenging hand-object interaction scenarios, which can generalize well to never seen real-world dataset and runs at 9 FPS.
(2) We propose the first point cloud-based hand pose tracking network, HandTrackNet, along with a novel hand canonicalization module, which together outperforms previous hand pose tracking or single-frame estimation methods.
(3) We synthesize a large-scale diverse simulated dataset for hand-object interaction that features realistic depth sensor noise and thus enables a direct Sim2Real generalization. We will release the data and the code upon acceptance.
\vspace{-3mm}
\section{Related Works}
\vspace{-3mm}
\subsection{3D Hand Pose Estimation and Tracking}
\vspace{-3mm}
Existing works on 3D hand pose estimation either learn representations in a 2D-3D paradigm, or directly operate on 3D input. In the 2D-3D paradigm, CNNs are used to process depth images directly, and predict 2D heatmaps of hand joints \cite{tompson2014real,guo2017region}, or regress 3D joint locations \cite{oberweger2015hands,madadi2017end,oberweger2017deepprior++,guo2017towards,guo2017region,zhang2019pixel}. A2J \cite{xiong2019a2j}and its following work \cite{virtulview} instead leverage an effective anchor-to-joint regression. RGB-based methods also use 2D CNNs, but heavily rely on 3D hand shape priors, which lift 2D heatmap predictions into 3D \cite{zhou2016model,zimmermann2017learning,spurr2018cross}, or predict parameters of MANO hand model \cite{baek2019pushing,boukhayma20193d,kulon2020weakly,romero2017embodied} and dense hand UV maps \cite{chen2021i2uv}. Instead of learning in 2D domain, depth images can be transformed into 3D space, either voxels \cite{ge20173d,moon2018v2v}, point clouds \cite{ge2018hand,li2019point,cheng2021handfoldingnet}, or T-SDF volume \cite{ge20173d}, in which the 3D information could be well preserved and learnt. In this work, we follow the recent trends to use 3D point clouds as input and perform 3D joints regression in 3D space.

Early works approach free hand pose tracking with hand-crafted features and post optimization \cite{oikonomidis2011efficient,qian2014realtime,tagliasacchi2015robust,taylor2016efficient} based on depth images by exploring articulation priors. When objects are present, they are usually considered as challenging occlusions like in \cite{mueller2017real}. Another line of work focuses on hand pose tracking from monocular videos \cite{mueller2018ganerated,wang2020rgb2hands,han2020megatrack}. 
Our method considers objects and hands at the same time and only relies on depth input.
\vspace{-3mm}

\subsection{Object Pose Estimation and Tracking}
\vspace{-3mm}

Historically, works on object pose estimation have focused on instance-level setting \cite{hinterstoier2012model,cao2016realtime,kendall2015posenet,xiang2017posecnn,wang2019densefusion,peng2019pvnet,he2020pvn3d,he2021ffb6d} which only deals with known object instances. 
\cite{wang2019normalized} extends object pose estimation to a category-level setting, where poses can be estimated for novel object instances from known categories. 
Recent works \cite{chen2020learning,chen2020category,tian2020shape,lee2021category,lin2021dualposenet,chen2021fsnet,huang2021stable} improve category-level pose estimation by leveraging shape synthesis \cite{chen2020category,chen2020learning,tian2020shape}, pose consistency \cite{lin2021dualposenet}, and geometrically stable patches \cite{huang2021stable}, as well as extending to RGB-only inputs \cite{lee2021category} and articulated objects \cite{li2020categorylevel}.

Besides single frame pose estimation, many recent works focus on the temporal tracking of object poses. Instance-level object pose tracking approaches include optimization \cite{pauwels2015simtrack,tan2015versatile,zhong2018robust,tjaden2019region}, filtering \cite{wuthrich2013probabilistic,choi2013rgb,krull20146,issac2016depth,deng2019poserbpf}, and direct regression of inter-frame pose change \cite{wen2020se}. 
Recent works on category-level object pose tracking can emerge category-level keypoints \cite{wang20196} without known CAD model in testing, refining coarse pose from keypoint registration by pose graph optimization \cite{wen2021bundletrack}, and learning inter-frame pose change from canonicalized point clouds \cite{weng2021captra}. Our method goes beyond category-level pose to novel object instances from both seen and unseen categories. 

\vspace{-3mm}
\subsection{Hand Object Interaction}
\vspace{-3mm}

Joint pose estimation/tracking and reconstruction of hand and objects during interaction has attracted much attention for its wide applications in virtual/augmented reality, teleoperation, and imitation learning \cite{hrst2011gesture,handa2020dexpilot,garciahernando2020physics,qin2021dexmv}. 
A number of datasets \cite{garcia2018first,hasson2019obman,hampali2020honnotate,hampali2020ho3dv2,taheri2020grab,brahmbhatt2020contactpose,chao2021dexycb} with both hand and object annotations have been developed to facilitate this line of research. Many works focus on single frame pose estimation from an RGB \cite{tekin2019h+o,hasson2019obman,kokic2019learning,hasson2020leveraging,doosti2020hopenet,chen2021joint,liu2021semi,karunratanakul2020grasping,yang2021cpf,zhang2021coarse,zhuang2021joint,cao2020reconstructing} or depth \cite{oberweger2019generalized} image, leveraging the interaction between the hand and the object by introducing feature fusion modules for joint prediction \cite{chen2021joint,liu2021semi,zhang2021coarse} and developing hand-object contact models \cite{hasson2019obman,karunratanakul2020grasping,grady2021contactopt,yang2021cpf,zhuang2021joint} to regularize poses. We similarly adopt penetration and attraction terms to explicitly model the hand-object interaction.

More related to our work are approaches that perform hand-object tracking over time. Early works use multi-view input to compensate for occlusions during interaction \cite{wang2013video,oikonomidis2011full}. More recent works take single-view RGB-D \cite{kyriazis2013physically,panteleris20153d,sridhar2016real,tsoli2018joint}, RGB-only \cite{hasson21homan}, or depth-only \cite{zhang2021single} video as input and leverage the physical contact constraints to reconstruct plausible poses. While most of them require a known object model or a shape template, \cite{panteleris20153d,zhang2021single} can deal with unknown objects. During tracking, they fuse depth observations of the object in a canonical frame to progressively reconstruct the object model, which is limited to the visible surface. Our method also handles unknown objects but reconstructs the complete object from the very first frame. In addition, while \cite{zhang2021single} relies on user-performed sequence collected in reality and VR, we only require synthetic training data. 


\vspace{-2mm}
\section{Methods}

\vspace{-2mm}
\subsection{Overview}
\vspace{-3mm}
\label{sec:prob_form}

\paragraph{Notations} In this paper, we deploy the following notations: 
right subscripts $(\cdot)_t$ represent quantities at the time step $t$; and left $_{C}(\cdot)$, $_{H}(\cdot)$, $_{O}(\cdot)$ denotes variables in camera space, canonical hand space (with zero global translation and identity rotation), and canonical object spaces, respectively. 

\vspace{-3mm}
\paragraph{Problem setting} Given a live stream of segmented hand and object point clouds$\{_C\PC^{hand}, _C\PC^{obj}\}_t$, along with initial hand joint positions $_C\J_{init}$ and initial object pose $\{\R_{init}^{obj}, \T_{init}^{obj}\}$ at frame $0$, our aim is to recover canonical object shape $_O\M^{obj}$ and its pose $\{\R_{t}^{obj}, \T_{t}^{obj}\}$, along with hand joint positions $_C\J_t$ and hand reconstruction, including canonical hand shape $_H\M_t^{hand}$
and global hand pose $\{\R_{t}^{hand}, \T_{t}^{hand}\}$, for all the following frames $t$ in an online manner.
Note that the object instance under tracking can be novel, but needs to come from the training categories of our object reconstruction model 
, or from categories whose geometry are similar to the training categories. For hand, we only focus on the right hand here, but our method is also suitable to the left hand.

\begin{figure}[t]
\centering
\begin{overpic}
[width=0.9\linewidth]
{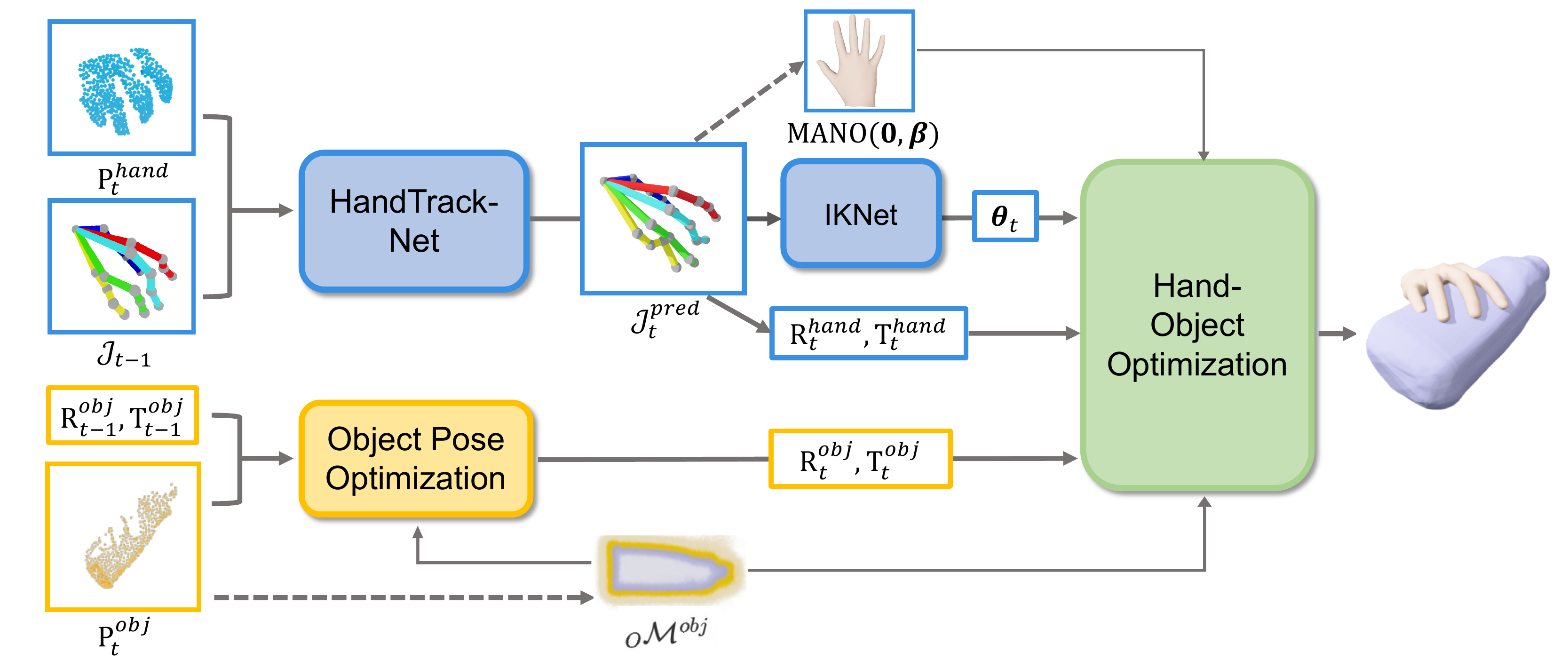}

\end{overpic}
\caption{
\textbf{The full pipeline.} At frame $0$, We initialize the object shape $_O\M^{obj}$ represented in signed distance function (SDF) and the hand shape code $\boldsymbol{\beta}$ for the parametric model MANO, as shown in the dotted line. In the following tracking phase, at each frame $t$, we first separately estimate the object pose $\{\R_{t}^{obj}, \T_{t}^{obj}\}$ and hand pose $\{\R_{t}^{hand}, \T_{t}^{hand}, \boldsymbol{\theta}_{t}\}$, and further refine the hand pose by taking hand-object interaction into consideration.
}
\label{fig:plot-pipeline}
\vspace{-3mm}
\end{figure}

\textbf{Pipeline overview.} 
The full pipeline is shown in Fig.\ref{fig:plot-pipeline}.
Prior to tracking, we need an initialization phase to estimate the object shape $_O\M^{obj}$ represented in SDF (Sec \ref{sec: obj track}) and the hand shape code $\boldsymbol{\beta}$ for the template-based parametric hand model MANO (Sec \ref{sec:hand model}) from the first frame.

Then, for object pose tracking, we simply perform a depth-to-SDF conformation based optimization \cite{zhang21rosefusion}.
For hand tracking and reconstruction, we combine neural network regression and optimization approaches: at each frame $t$, we first leverage our point cloud based neural network, HandTrackNet, to update the hand joint positions $\J_t$ based on  $\J_{t-1}$ (Sec. \ref{sec: HandNet}); then, to recover the full geometry of hand, we convert $\J_t$ into MANO hand $_C\M_t^{hand}$ with the help of IKNet (Sec. \ref{sec:hand model}); finally, we refine the hand pose based on hand-object interaction priors to recover more physical plausible interaction.



\vspace{-3mm}
\subsection{Tracking Hand Joint Positions} 
\vspace{-2mm}

\begin{figure}[t]
\centering
\begin{overpic}
[width=\linewidth]
{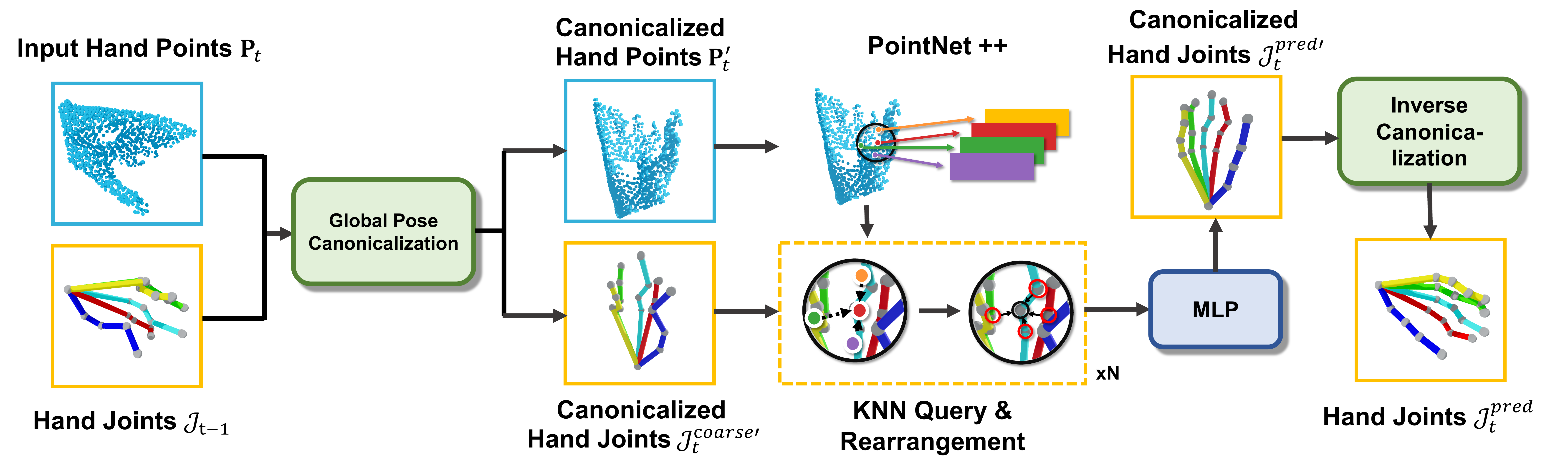}
\end{overpic}
\vspace{-3mm}
\caption{\textbf{The architecture of HandTrackNet.} HandTrackNet takes input the hand points $\PC_t$ from the current frame $t$ and the estimated hand joints $\J_{t-1}$ from the last frame, and perform global pose canonicalization to both of the two. Then it leverage PointNet++\cite{qi2017pointnet++} to extract features from canonicalized hand points $\PC_t'$ and use each joint $\J_t^{coarse'}$ to query and pass features, followed by a MLP to regress and update joint positions.}
\label{fig:plot-network}
\vspace{-3mm}
\end{figure}

\label{sec: HandNet}
In this section, we introduce our point cloud based neural network, HandTrackNet, for tracking hand joint positions. During hand object interaction, many hand joints are heavily occluded, making joint position regression very challenging and ambiguous. We thus devise HandTrackNet to leverage hand joint positions $\J_{t-1}$ from the previous frame prediction, which distinguishes this work from all previous single-frame hand pose estimation methods \cite{cheng2021handfoldingnet, ge2018hand} that directly regress $\J_t$ from the current observation $\PC_t^{hand}$. 
We leverage $\J_{t-1}$ in two ways: first, $\J_{t-1}$ can provide a rough global pose of the hand, which can be used to canonicalize the input hand point cloud $\PC_t$, to reduce the learning complexity and increase the generalizability of different hand global poses; second, the network only needs to regress the small changes for updating $\J_{t-1}$, which reduces the output space of the neural network and thus eases the regression. The whole architecture of HandTrackNet is shown in Figure \ref{fig:plot-network}.

\label{sec:canon}
\textbf{Hand pose canonicalization.} As pointed out by Hand PointNet~\cite{ge2018hand}, the large variations in hand global orientation bring in significant complexity and big challenges to 3D hand pose estimation. To reduce this visual complexity, ~\cite{ge2018hand} proposed a point cloud canonicalization method that canonicalizes the orientation of the hand point cloud using the oriented bounding box (OBB), whose $x,y,z$ axes are approximated by the PCA directions of input points.
As a pre-processing step, this orientation canonicalization is widely adopted by the following works for single hand pose estimation, \textit{e.g.} P2P\cite{ge2018point} and HandFoldingNet\cite{cheng2021handfoldingnet}, demonstrating consistent performance improvements.

However, in the scenario of hand-object interaction, the input hand point clouds can be heavily occluded and become very partial, making PCA less effective for estimating the hand orientation and thus nullifying the effect of orientation canonicalization. To overcome this issue,
we propose a simple yet novel method to estimate the 6D transformation between camera \textit{c.s.} and hand \textit{c.s.} from the joint positions $\J_{t-1}$ and then use it for canonicalization. 

We start from a common assumption \cite{yuan2017bighand2} for 3D hand pose fitting -- the wrist and 5 metacarpophalangeal (MCP) joints (where the finger bones meet the hand bones) move together and thus are fixed relative to each other. We denote this set of joints as $B$ and their spatial positions as $\J^{B} = \{\J^j~|~j\in B\}$. We can further define a function $CanonPose$ to solve the rigid transformation for canonicalization:
\begin{equation}
    CanonPose(\boldsymbol{\beta},~_C\J^B) \triangleq \arg\underset{\R, \T}{\min}\|_H\J^{B}_{MANO}(\boldsymbol{\beta})-\R^{-1}(_C\J^B-\T)\|^2
    \label{eq: canon}
\end{equation}
in which $_H\J_{MANO}(\boldsymbol{\beta})$ denotes the joint positions of the template-based parametric hand model MANO, introduced in Sec. \ref{sec:hand model}, with zero translation and identity rotation. Note that $_H\J_{MANO}^B$ is irrelevant to the pose vector $\boldsymbol{\theta}$ because the bending of the fingers won't influence the position of $B$. The shape code $\boldsymbol{\beta}$ is initialized at frame $0$. This argmin function can be analytically solved by SVD. 

Finally we can solve $\R^B_t, \T^B_t = CanonPose(\boldsymbol{\beta},~_C\J^B_{t-1})$ and use it to canonicalize the hand point cloud $~\PC_t' = (\R_{t}^{B})^{-1}(\PC_t-\T_{t}^{B})$.
Due to temporal continuity, $\PC_t'$ only has a small rotation and translation, which simplifies the network input and narrows down the regression output space, reducing the difficulty of regression learning.

\textbf{Network architecture.}  After canonicalizing $\PC_t$ to $\PC_t'$, a naive solution is using a PointNet++\cite{qi2017pointnet++} to directly regress the hand joint positions $\J_t'$. However, under the heavy self- and inter- occlusion, the positions of invisible joints may still be ambiguous.
To alleviate this issue and ensure the joints won't go arbitrarily far, we propose to leverage the estimated joint positions from the last frame to provide a coarse estimation $\J_t^{coarse'} = (\R_{t}^{B})^{-1}(\J_{t-1} + \bar{\PC}_{t} - \bar{\PC}_{t-1}
-\T_{t}^{B})$, which means we add the center shift between $\PC_t$ and $\PC_{t-1}$ to $\J_{t-1}$ and then transform it to the canonical frame. Now, to obtain $\J_t'$, the network will only need to regress a small movement for each joint and add them to $\J_t^{coarse'}$.

Our network first extracts the per-point feature of $\PC_{t}'$ using PointNet++. Then, for each hand joint of $\J_t^{coarse'}$, we find k nearest neighbors in $\PC_{t}'$, and aggregate those neighbors' features by a PointNet-based\cite{qi2017pointnet} structure to encode the local information of $\J_t^{coarse'}$. To further enlarge the receptive field and encode the global information about each joint, we add a rearrangement layer to communicate among joints by concatenating the feature of adjacent joints, which is inspired by HandFoldingNet\cite{cheng2021handfoldingnet}. Our rearrangement module is different from theirs because they only communicate among joints in the same finger but we also communicate with the neighbor finger. This design can provide more global information of a hand and avoid confusing the joints of different fingers, as shown in our supplementary. The above two steps, named KNN query and rearrangement, are repeated 
N times (we set N=2 in our paper, and we investigate different Ns in supp.) to improve the network expressivity. Finally, we use the feature of each hand joint to regress the relative movement $\J_t'-\J_t^{coarse'}$, and transform the $\J_t'$ back to the camera \textit{c.s.} using the inverse transformation of canonicalization.

\textbf{Training strategy}
\label{train detail}
To train HandTrackNet, we use single frame data and randomly perturb the ground truth hand pose by adding a Gaussian noise(we set std=2cm). Our training loss is designed as 
$\loss{} = \lambda_{joints}\|\J_{gt}-\J_{pred}\|_1 + \lambda_{rot}\|\R_{gt}-\R_{pred}\|_1 + \lambda_{trans}\|\T_{gt}-\T_{pred}\|_1$, where $\R_{pred}, \T_{pred} = CanonPose(\boldsymbol{\beta},~_C\J_{pred}^B$) and they are supervised to ensure a good canonicalization for the next frame in testing. For hyper-parameters, we set $\lambda_{joints}=10$ and $\lambda_{rot}=\lambda_{trans}=1$.

\vspace{-2mm}
\subsection{MANO Hand Reconstruction}
\vspace{-2mm}

During hand object interaction, the kinematic chain of a hand will always remain the same and only joint states may change. 
Although our HandTrackNet has taken the temporal continuity into account, there is no guarantee that the hand joints will always remain the same structure, which may lead to artifacts, \textit{e.g.} implausible joint positions and inconsistent bone length.
Furthermore, without a hand shape, joint positions alone are not sufficient to investigate the interaction between the hand and the object. Thus, we propose to reconstruct the hand shape $_{C}\M_t^{hand}$ at each frame. 

\label{sec:hand model}
\textbf{Parametric hand model.}  We use MANO \cite{MANO:SIGGRAPHASIA:2017}, a popular template-based parametric layer, as our hand model. It maps a shape vector $\boldsymbol{\beta}\in\real^{10}$ and a pose vector $\boldsymbol{\theta}\in\real^{45}$ to a mesh $_H\M^{hand}$ and the corresponding joint positions $_H\J$. Specifically, $\boldsymbol{\beta}$ is the coefficients of learned shape PCA bases, and $\boldsymbol{\theta}$ represents the rotation of 15 joints in axis-angle representation. 
See \cite{MANO:SIGGRAPHASIA:2017} for more details.

\textbf{Shape code initialization.} 
Since bone lengths (or the distances between two neighboring joints) keep unchanged regardless of the joint angle $\boldsymbol{\theta}$, we can optimize the shape code $\boldsymbol{\beta}$ by minimizing
\begin{equation}
    E(\boldsymbol{\beta}) = \| l(_C\J_{pred}) - l(_H\J_{MANO}(\mathbf{0}, \boldsymbol{\beta})) \|_1
\end{equation}
in which $l(\J)\in\real^{15}$ means the length of 15 finger bones of joints $\J$, and $_C\J_{pred}$ is the prediction of HandTrackNet. In our paper we only estimate the shape code $\boldsymbol{\beta}$ at frame $0$ for simplicity and consistency. In Supp. we also discuss about updating $\boldsymbol{\beta}$ during tracking.

\textbf{Inverse kinematic network (IKNet).} To obtain joint angles $\boldsymbol{\theta}_t$, we devise a simple MLP that takes input joint position $\J'_t$ and output joint angles $\boldsymbol{\theta}_t$, which indeed does the job of inverse kinematics. Such a network can achieve a very high speed while also remain good performance, which suits our need well to serve as a better initialization than $\boldsymbol{\theta}_{t-1}$ for the following hand-object reasoning. 

The difference between ours and \cite{zhou2020monocular} is that we use the same canonicalization methods as HandTrackNet, which ensure the generalization ability (see our supp. for experiments). Specifically, we solve the global hand pose $\R_t^{hand}, \T_t^{hand}=CanonPose(\boldsymbol{\beta},~_C\J^B_{pred})$
 and canonicalize the joint position $\J'_t=(\R^{hand}_t)^{-1}(\J_t-\T^{hand}_t)$, which serves as the direct input to the MLP. Finally, we can reconstuct the hand shape $_C\M^{hand}_t=\R^{hand}_t~_H\M_{MANO}(\boldsymbol{\theta}_t, \boldsymbol{\beta})+\T^{hand}_t$.

\subsection{Object Reconstruction and Pose Tracking}
\label{sec: obj track}
In contrast to \cite{zhang2021single} who use a fusion-based method to reconstruct the novel object and may fail to recover the full geometry if some places are always occluded, leading to further failure in the following hand object reasoning, we leverage category shape prior to recover the full object shape at the first frame. Specifically, we propose to first estimate the signed distance function (SDF) of the object from the first frame, and then minimize a depth-to-SDF conformance~\cite{zhang21rosefusion} based energy function to optimize pose. SDF \cite{park2019deepsdf} is an implicit 3D representation, which maps a spatial coordinate $\x$ to its signed distance $s$ to the closest surface, \textit{i.e.} $SDF(\x)=s~:~\x\in\real^3, s\in\real$.

\paragraph{Shape initialization.} We train a DeepSDF \cite{park2019deepsdf} using synthetic objects in ShapeNet\cite{chang2015shapenet}. During testing, at frame 0, given the real partial point cloud of an object, we first use the given initial object pose to canonicalize this point cloud and then optimize the shape code of DeepSDF by minimizing the mean square loss of the SDF values of the depth points. Following \cite{park2019deepsdf}, we also use a L2 regularization on the shape code to alleviate the affect of the noisy real point clouds.  

\paragraph{Object pose optimization} At frame $t$, given $_C\PC_t^{obj}$, together with the signed distance field $\psi:\real^3 \rightarrow \real$ of the reconstructed object shape $_O\M^{obj}$, we optimize $\{\R_{t}^{obj}, \T_{t}^{obj}\}$ to minimize

\begin{equation}
    E(\R_{t}^{obj}, \T_{t}^{obj}) = \sum_{\x\in _C\PC_t} \left|\psi[(\R_t^{obj})^{-1}(\x-\T_t^{obj})]\right|
\end{equation}
For the choice of optimizer, we empirically find the gradient-based optimizer, \textit{e.g.} Adam and LM, both suffer from a low speed. Therefore, we use a recently proposed gradient-free optimization method \cite{zhang21rosefusion} based on random optimization. By taking the advantage of its highly parallelized framework, our method can efficiently converge to a good local optimum at a frame rate of 29 FPS under PyTorch. Please see our Supp. for more details about the comparison of different optimizers.

\paragraph{Shape update} In experiments, we find that our method can already generalize well to the object from a novel categories, \textit{e.g.} trained on bottle and tested on box and can. To further mitigate the shape gap cross category and enhance the generalization ability, we also combine our method with those fusion-based method to further update the object shape during tracking. Please see our supp. for more details. 

\vspace{-3mm}
\subsection{Hand-Object Optimization}
\label{sec:ho opt}
\vspace{-2mm}
Separately estimating the pose of hand and object often suffers from the problem of unrealistic hand-object interactions, as discussed in lots of related works \cite{hasson2019obman, hasson21homan}. Based on their works, we adopt the energy terms to refine the hand pose $\{\boldsymbol{\theta}, \R^{hand}, \T^{hand}\}$ by optimization at each frame.

Here we focus on using objects to help hand pose because the hand is a high DoF articulated object and the poses of those invisible fingers are very ambiguous, which is hard for HandTrackNet to estimate correctly but can be refined using certain physical constraints with the object. In contrast, the freedom and ambiguity of object pose are much smaller, which relieves the need from hand. In our experiment (see our supp.), the error of object pose will even increase if we jointly optimize the hand pose and object pose. In the following, we will briefly introduce the energies used in this stage.

\textbf{Penetration and attraction energy.} Two commonly used energies for physical constraint, as proposed in \cite{hasson2019obman}, are to punish hand-object penetration and encourage the contact between certain hand regions and object. However, unlike previous works \cite{hasson2019obman, cao2020reconstructing} which only focus on the static in-contact hand object interactions, for general hand object interaction we should also consider when the hand and object are not in contact. We propose a simple strategy to decide whether to use the attraction energy by the magnitude of penetration energy. See our supplementary for details.

\textbf{Joint position energy.} To maintain the accuracy of joint positions, we enforce an energy to punish the L2 distance between $\R^{hand}~ _H\J_{MANO}(\boldsymbol{\theta})+\T^{hand}$ and $_C\J_{pred}$. Since the invisible joint positions predicted by HandTrackNet may suffer from ambiguity and have a large error, this energy is only used for the visible joints. The visibility of joint $i$ is judged by whether the distance of $\J_i^{pred}$ to $\PC^{hand}$ is larger than a threshold $\lambda$. we set $\lambda=2$cm for all of our experiments.

\textbf{Silhouette energy.} To ensure the consistency between our prediction and the observation, we project the reconstructed hand shape to the image plane and add a silhouette loss to punish the points which are outside the silhouette of hand and object.
\vspace{-3mm}
\subsection{Training Data Generation via Depth Simulation}
\vspace{-2mm}
To avoid using expensive real world annotation for training HandTrackNet, we propose to synthesize a large-scale point cloud dataset for hand and object interaction which we call the SimGrasp dataset.
Our aim is to generate realistic point cloud sequences for diverse hand object interactions with randomized object shapes, hand shapes, grasping poses, and motion trajectories. We introduce the details of our data generation method in supplementary material. As a result, SimGrasp contains 666 different object instances and 100 hand shapes, which forms 1810 videos with 100 frames per video. 

To mitigate the domain gap between synthetic depths and real depths, 
we re-implement a structure light based depth sensor simulator, DDS\cite{planche2021physics}, for depth image synthesis so that our simulated point clouds can capture realistic noises and artifacts. 
This depth simulation technique along with our fully domain randomized hand object interaction synthesis enables a direct Sim2Real transfer and allows us to test our HandTrackNet on real data in the wild.

\section{Experiments}

\subsection{Datasets and Metrics}
\label{metric}

To evaluate the performance of our method on the in-the-wild data, we use two popular real world hand-object interaction datasets, HO3D\cite{hampali2020honnotate} and DexYCB \cite{chao2021dexycb}, whose hands, object instances, and motion trajectories are never seen during training. 
HO3D dataset is a widely used RGB-D video dataset, and DexYCB dataset is a more challenging dataset because of the larger diversity and faster motion. For a fair comparison with previous methods, we train all the methods purely on SimGrasp and directly test HO3D and DexYCB. Please see our supp. for more details about these three datasets.

 We report the following metrics to evaluate the results. 1) \textbf{MPJPE}: we use the mean per joint position error for evaluating the hand pose. 2) \textbf{Penetration depth (PD)}: following \cite{yang2021cpf}, we report maximum penetration depth between hand and object to measure the physical plausibility of the reconstructed hand and object. 3) \textbf{Disjointedness distance (DD)}: following \cite{yang2021cpf}, we report the mean distance of hand vertices in 5 finger regions to their closest object vertices in frames when the ground truth hand and object are in contact.
 4) \textbf{5$^\circ$5cm, 10$^\circ$10cm}: we follow \cite{weng2021captra} to report the 5$^\circ$5cm accuracy of object pose, which means the percentage of pose predictions with rotation error < 5$^\circ$ and translation error < 5cm. 10$^\circ$10cm is defined similarly.
 5) \textbf{Chamfer distance (CD)}: to measure the error of the posed object geometry during tracking, we report an adapted chamfer distance $CD(_C\M^{obj}_{pred},~_C\M^{obj}_{gt})$, in which $CD(\M_1, \M_2)=\frac{1}{|\M_{1}|}\sum_{\x\in \M_{1}}\min_{\y\in \M_{2}}\|\x-\y\|_2+\frac{1}{|\M_{2}|}\sum_{\y\in \M_{2}}\min_{\x\in \M_{1}}\|\x-\y\|_2$. 

\begin{table}[htbp]
\caption{\textbf{Hand pose tracking.} All the methods are trained and validated on SimGrasp, and directly tested on real world dataset. '-' denotes the method doesn't reconstruct the hand shape thus can't be evaluated for those metrics. }
\label{table:hand pose}
    \resizebox{\columnwidth}{!}{
        \begin{tabular}{l|ccc||ccc|ccc}
        \toprule
        &  \multicolumn{3}{c||}{Simulated Validation}& \multicolumn{6}{c}{Real World Testing} \\
       \midrule
        & \multicolumn{3}{c||}{SimGrasp} & \multicolumn{3}{c}{HO3D} & \multicolumn{3}{|c}{DexYCB}\\
         & MPJPE (cm) & PD (cm) & DD (cm)& MPJPE (cm) & PD (cm) & DD (cm)& MPJPE (cm) & PD (cm) & DD (cm)\\
        \midrule
        Forth~\cite{oikonomidis2011efficient}&2.51&-&-&4.04&-&-&4.19&-&-\\
        HandFoldingNet~\cite{cheng2021handfoldingnet}&1.22&-&-&2.93&-&-&3.78&-&-\\
        A2J~\cite{xiong2019a2j}& 0.95&-&-&4.03&-&-&3.52&-&- \\
        VirtualView~\cite{virtulview}&0.91&-&-&2.73&-&-&3.05&-&-\\
        \midrule
        HandTrackNet&\textbf{0.84}&-&-&2.11&-&-&2.75&-&-\\
        Ours w/o hand-obj opt. & 0.96& 1.44 & 1.18& 2.34 & 1.42& 1.54& 2.99& 1.82& 1.81\\
        Ours (w/ hand-obj opt.) &0.90& \textbf{1.31}& \textbf{1.11}&\textbf{2.06}& \textbf{1.08}& \textbf{1.21}&\textbf{2.69}&\textbf{1.48}& \textbf{1.45}\\
        \bottomrule
        \end{tabular}
    }
\end{table}

\vspace{-3mm}
\subsection{Results of Hand Pose Tracking and Reconstruction}
\vspace{-2mm}
\textbf{Hand joint positions.} In Table \ref{table:hand pose}, we compare our approach against state-of-the-art single-frame point cloud-based method HandFoldingNet~\cite{cheng2021handfoldingnet}, depth-based method VirtualView~\cite{virtulview} and A2J~\cite{xiong2019a2j}, as well as an optimization-based tracking method Forth~\cite{oikonomidis2011efficient}. All the methods are trained on our SimGrasp and tested on real datasets without any fineture. Note that, though we achieve relatively minor improvement on the simulated dataset, our method shows very promising results when testing on real world datasets, which demonstrates the generalization ability and robustness to domain gap.

\textbf{Hand reconstruction and optimization.}
In addition to predicting hand joint positions, our method also reconstructs the hand mesh.
We observe that although the joint error actually increases after using the inverse kinematic network to convert joint positions into the MANO hand mesh, the following hand object optimization module significantly reduces this error and achieves the best performance among all methods on the two real datasets. For the penetration depth and the disjointedness distance, they benefit from optimization as well, dropping by about 3mm on both real datasets. Apart from quantitative results, the visual quality of our reconstruction is also improved (see our Supp.).

\subsection{Object Pose Tracking and Reconstruction}
Since the object mesh is unknown for the sequences in the wild, instance-level object pose tracking methods aren't suitable for this scenario. Therefore we choose the state-of-the-art point cloud based category-level object pose tracking method CAPTRA\cite{weng2021captra} for a comparison. However, category-level methods will also meet the situation where the testing object category shares no overlap with the training data, which creates a new challenge for the generalization ability of the methods.

In our experiments, for SimGrasp both of the two methods are trained and evaluated in the same category. As for HO3D and DexYCB, these two methods are trained on the most similar category in SimGrasp for evaluation. Note that since CAPTRA don't reconstruct the object, we use their pose to transform our reconstruction $_O\M^{obj}$ to $_C\M^{obj}$ for evaluation. See supp. for more details. 

The results are reported in Table \ref{table:object pose}. We can see that though CAPTRA achieves an impressing result on SimGrasp, it fails to generalize in most cases. For HO3D, CAPTRA can do well for the object pose of the boxes, but it has a larger Chamfer distance error than ours and fails in the bottles which have a larger inter-category gap to the training objects (\textit{i.e.} car). For DexYCB, the distortion of the point cloud caused by fast motion enlarges the data distribution gap, resulting in the bad performance of CAPTRA. In contrast, our method generalizes well in all the settings. Furthermore, our tracking system can also handle the sequences of the hand in interaction with the objects from a very different category (\textit{e.g.} scissors and drills) to our training category of DeepSDF \cite{park2019deepsdf}, if the ground truth object mesh is known during testing, as shown in the 'others' in Table $\ref{table:object pose}$.

\begin{table}[htbp]
    \caption{\textbf{Object pose tracking.} '$^\dagger$' denotes that the object category has symmetry and its rotation is evaluated by the error of symmetric axis direction. '*' denotes that the ground truth object mesh is given. See the supplementary for the instances contained in each category of HO3D and DexYCB.}
    \resizebox{\columnwidth}{!}{
        \begin{tabular}{cc|ccc|ccc|ccccc}
        \toprule 
        & & \multicolumn{3}{c}{Simulated Validation}& \multicolumn{8}{|c}{Real World Testing} \\
        \midrule
        && \multicolumn{3}{c}{SimGrasp} & 
        \multicolumn{3}{|c|}{HO3D} &
        \multicolumn{5}{c}{DexYCB} \\
        & & bottle$^\dagger$ & car & bowl$^\dagger$ & bottle & box & others* &bottle & bowl$^\dagger$ & box & can$^\dagger$ & others* \\
        \midrule
          & 5$^\circ$5cm($\%$) $\uparrow$&  
          \textbf{84.6}& \textbf{87.9}&
          \textbf{56.7}&15.7&\textbf{67.3}&-& 16.9&23.5&23.6&22.2&-\\
        CAPTRA& 10$^\circ$10cm($\%$) $\uparrow$ & \textbf{95.5}&\textbf{96.4}&
        \textbf{80.5}&48.5&\textbf{91.9}&-& 42.3&54.1&46.5&55.0&-\\
        \cite{weng2021captra}& CD(cm) $\downarrow$ &  \textbf{1.84}&\textbf{2.21}&
        \textbf{2.17}&2.84&2.22&-& 3.97&4.97&3.83&4.93&-\\
        \midrule
        & 5$^\circ$5cm($\%$) $\uparrow$ & 80.1 & 76.0 &49.1 & 
        \textbf{43.7} &54.5 & \textbf{57.0}
         & \textbf{38.5} &\textbf{33.3} 
        &\textbf{31.2}&\textbf{35.7}
        & \textbf{36.8}\\
        Ours & 10$^\circ$10cm($\%$) $\uparrow$& 90.1 & 88.1 & 66.5
        &\textbf{68.9} & 84.6&\textbf{82.3}
        &\textbf{64.1}&\textbf{58.4}
         &\textbf{52.0}&\textbf{59.0}&\textbf{58.5}\\
        & CD(cm) $\downarrow$&  2.15 & 2.28 & \textbf{2.17} & \textbf{2.06} & \textbf{1.74} &\textbf{1.20}
        & \textbf{2.46}&\textbf{4.71}
        &\textbf{2.82}&\textbf{2.32}&\textbf{2.30}\\
        \bottomrule
        \end{tabular}
    }
    \label{table:object pose}
   
\end{table}

\subsection{Ablation Study}
\begin{figure}[htbp]
\caption{Left: Ablation study on HandTrackNet. Right: Robust analysis. We show the mean joint error over videos along time on HO3D dataset with different initialization noise of $\J_{init}$. }
\label{table:ablation}

\begin{minipage}[h]{0.5\columnwidth}
    \resizebox{\columnwidth}{!}{
        \begin{tabular}{l|ccc}
        \toprule
        MPJPE(cm) & SimGrasp & HO3D & DexYCB\\
        \midrule
        \textit{w/o stereo depth simulation} & 2.53 & 2.49 & 3.68\\
        \textit{w/o canonicalization} & 1.30 & 4.73 & 10.9\\
        \textit{use OBB for canon. } & 2.83 & 4.95 & 4.00 \\
        \textit{w/o $\J_{t-1}$} & 0.93 & 2.46 & 2.95\\
        \textit{w/o rearrange intra-fingers} &1.16 & 2.69 &2.82\\
        \midrule
        HandTrackNet&\textbf{0.84}&\textbf{2.11}&\textbf{2.75}\\
        \bottomrule
        \end{tabular}
    }
\end{minipage}
\begin{minipage}[h]{0.5\columnwidth}
\includegraphics[width=\columnwidth]{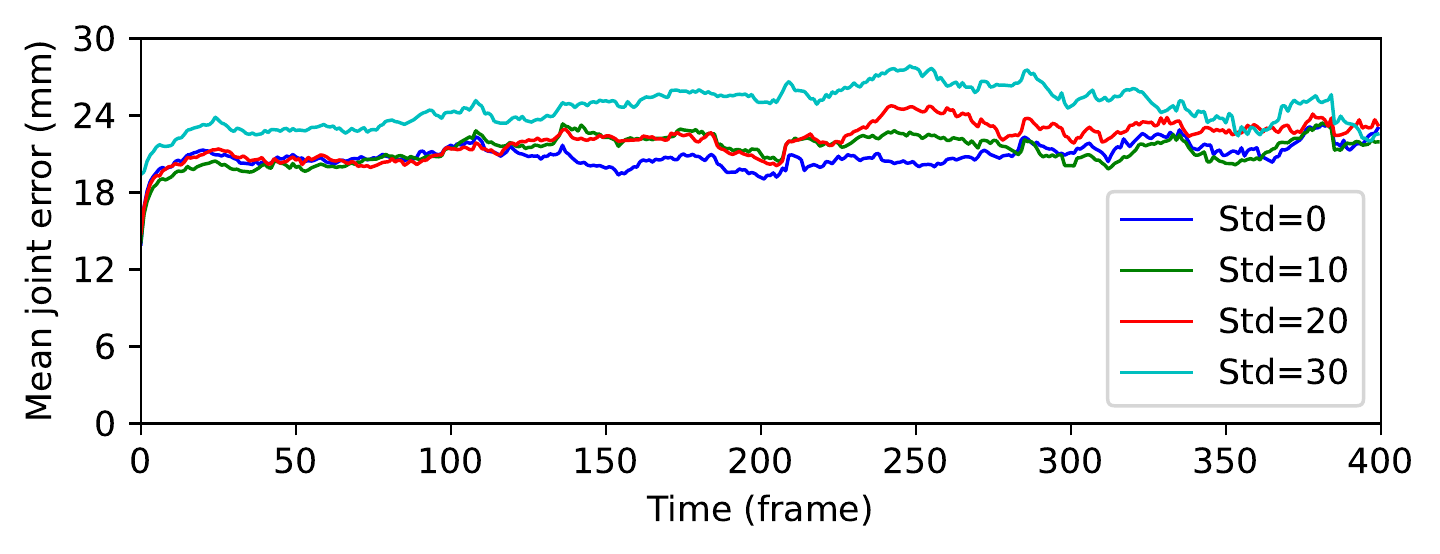}
\end{minipage}

\end{figure}

We first verify our stereo-based depth simulation. By replacing the simulated depth with perfect depth, we generate a clean synthetic dataset for training. Figure \ref{table:ablation} shows that the performance drops a lot on all three datasets, showing that our simulated sensor greatly reduces the Sim2Real gap. 

Then, we examine our canonicalization module. If we simply remove this module or replace it with the pre-processing method OBB, the performance will drop a lot, especially on real dataset, indicating that the canonicalization module greatly improves the generalization ability.  

We also show the importance of the last frame's joint positions $\J_{t-1}$ by replacing $\J^{coarse'}$ with a coarse estimation predicted by the bottleneck feature of PointNet++. We observe that the performance drops mainly comes from the temporal inconsistency under occlusion. Finally, we test the rearrangement module used in HandFoldingNet, which doesn't communicate joint features among different fingers. Without the information from the neighbor fingers, the network becomes easily confused by the joints on the different fingers during tracking. Some typical failure cases are shown in our supplementary materials.

\subsection{Robustness and Speed}
\textbf{Robustness.} We analyze the robustness of our final tracking methods against the noisy inputs $\J_{init}$ on HO3D. The initial joint position errors are randomly sampled from Gaussian distributions. We show the results with different standard deviations from 10 to 40mm in Figure \ref{table:ablation}. We can see that our method is robust to the initialization with some reasonable noise and can keep the joint error at a low level along time. We leave to future study for further increasing the tolerance of initialization errors.

\textbf{Tracking speed.} In the tracking phase, our full pipeline runs at 9 FPS in PyTorch without any C++ or CUDA acceleration tricks on an RTX 2080 Ti GPU. The HandTrackNet and object tracking module runs at 26 and 29 FPS respectively, which can be paralleled. The IKNet can reach 50FPS and the final hand-object optimization runs at 19FPS. In the intialization phase, the object takes 1 second and the hand only takes 0.15 second.

\vspace{-4mm}
\section{Conclusions} 
\label{conclusion}
\vspace{-3mm}
In this paper, we present a hand-object tracking and reconstruction system with powerful generalization capability. 
 Through extensive experiments, our method shows state-of-the-art performance on real world datasets. In the future, we would like to update the object shape according to the newly observed point cloud during tracking and also speed up the system by CUDA acceleration. Besides, our method could be further improved by turning the full pipeline into an end-to-end manner.

%
%
\bibliographystyle{splncs04}
\bibliography{egbib}

\clearpage

\appendix

\begin{abstract}
    In this supplementary material, we first provide more details about the rearrangement layer in our HandTrackNet in Sec. \ref{sec: handnet details}, and the energy term for hand-object reasoning in \ref{sec: ho energy}. Then we introduce the gradient-free optimizer in Sec. \ref{sec: particle swarm}, including the experiments comparing to other optimizers. Next, we discuss about the hand shape update in \ref{sec: hand shape update} and object shape update in Sec. \ref{sec: object shape update}, along with the experiments and analysis. We introduce more details about our SimGrasp datasets and real world testing datasets in Sec. \ref{sec: simgrasp} and Sec. \ref{sec: testing datasets}, respectively. Then we add more ablation studies in Sec. \ref{sec: more ablation}, and show some visualizations in Sec. \ref{sec:result_vis}. Last but not least, we discuss about the limitation and the social impact of our paper in Sec. \ref{sec: limit}.
\end{abstract}

\section{Rearrangement for HandTrackNet}

\label{sec: handnet details}



\begin{figure}[htbp]
    \centering
    \includegraphics[width=0.18\columnwidth]{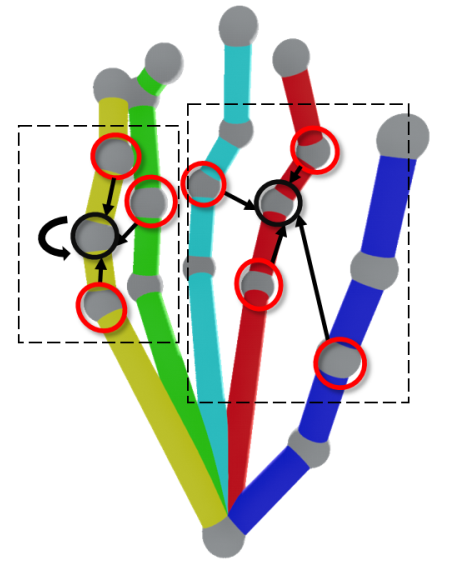}
    \caption{\textbf{Illustration for rearrangement.} For example, the joint on the red finger in the black circle will communicate with four neighbors. The joint on the yellow finger in the black circle will use the feature of itself to represent the left neighbor.}
    \label{fig:rearrange}
\end{figure}

Different from the one in HandFoldingNet\cite{cheng2021handfoldingnet}, for each joint we concatenate the feature of its four neighboring joints from the top, bottom, left and right, as shown in Figure \ref{fig:rearrange}. If one joint has no neighbors on one side, it will use the feature of itself to represent the missing neighbors.

\section{Energy Terms for Hand-Object Optimization}
\label{sec: ho energy}
At each frame, we optimize $\{ \boldsymbol{\theta}, \R^{hand}, \T^{hand}\}$ to refine the estimated hand. We also use the gradient-free (or searching-based) optimization introduced in Sec. \ref{sec: particle swarm}. To reduce the dimension of search space, we sample $\Delta\boldsymbol{\theta} \in \real^{10}$, which corresponds to the first ten pose PCA components of MANO.

\paragraph{Penetration and attraction loss} It is not trivial to adopt this two terms for general hand object interaction. Unlike previous works \cite{hasson2019obman, cao2020reconstructing} which only focus on the static in-contact hand object interactions, in an unconstrained video we also need to deal with the situation when the hand and object have no contact, \textit{e.g.} during the pre-grasping stage, which means that we can't use the attraction energy all the time. 

We propose a simple strategy to decide whether to use the attraction energy by computing the maximum penetration: if the maximum penetration is smaller than a threshold, we think the hand has not reached the object and doesn't need such an attraction term; otherwise, attraction energy is needed to pull over the distant fingers. 

This strategy seems to have a very bad local minima, where the estimated hand and object may have a large translation error and thus are far away from each other. In this situation, the attraction term can't be used to pull over the hand and object using our strategy. However, since our coarse poses of hand and object are estimated from point clouds and won't have a large translation error like RGB-based method \cite{cao2020reconstructing}, this simple strategy will not suffer from this bad local minima. 

The energy terms are 
\begin{equation}
    E_{penetr}(\boldsymbol{\theta}, \R^{hand}, \T^{hand}) = \underset{\x \in _C\M^{hand}}{\max} -\mathbbm{1}_{\psi(\x)<0}\psi(\x)
\end{equation}

\begin{equation}
 E_{attr}(\boldsymbol{\theta}, \R^{hand}, \T^{hand}) = \underset{\x \in CR} {\Sigma}\mathbbm{1}_{\psi(\x)>0} \psi(\x)
\end{equation}

in which $_C\M^{hand}=\R^{hand}~_H\M^{MANO}(\boldsymbol{\theta}, \boldsymbol{\beta})+\T^{hand}$, and $\psi(\x)$ means the signed distance of the estimated object $_C\M^{obj}$, and $CR$ is the pre-defined contact regions on hand finger tips following \cite{hasson2019obman}. 

\paragraph{Joint position loss}
\begin{equation}
     E_{joint}(\boldsymbol{\theta}, \R^{hand}, \T^{hand}) = \| _C\J^{pred} - (\R^{hand}~ _H\J^{MANO}(\boldsymbol{\theta}, \boldsymbol{\beta})+\T^{hand})\|_1
\end{equation}

\paragraph{Silhouette loss}
\begin{equation}
    E_{sil}(\boldsymbol{\theta}, \R^{hand}, \T^{hand})=\underset{\x \in _C\M^{hand}}{\Sigma}\mathbbm{1}_{project(\x)\in S}
\end{equation}
in which $S=Silhouette^{hand}\cup Silhouette^{obj}$ and $_C\M^{hand}=\R^{hand}~_H\M^{MANO}(\boldsymbol{\theta}, \boldsymbol{\beta})+\T^{hand}$. Since we use gradient-free optimization, this energy doesn't have to be differentiable.

\section{Gradient-free Optimizer and Ablation Study}
\label{sec: particle swarm}
\subsection{Particle Swarm Algorithm}

We use a recently proposed gradient-free (or searching-based) optimization, particle swarm algorithm\cite{zhang21rosefusion}, in our paper. The main idea is to pre-sample a great number of particles from $\mathcal{N}(0, 1)$ and in each iteration use a search step $\mathbf{s}$ to rescale those particles, which are then served as the possible change of variables to be optimized. Next, compute the energy for each particle and use those good particles to update the search step and variables, until convergence. Please see \cite{zhang21rosefusion} for details.

In our case, to avoid the heavy computational cost of inferencing the DeepSDF decoder for each time, we pre-compute and save the signed distance of the object in discrete grids. In such a way we only need to query and interpolate when computing the energy, greatly speeding up the computation. Note that we don't need to repeat computing the SDF grids in each frame unless the object shape is updated.

\subsection{Comparison to Different Optimizers}
 We compare different gradient-based optimizers, Adam and LM, to our gradient-free optimizer for object pose tracking in Table \ref{tab:optimizer}. For LM, we use the most recent work DeepLM\cite{huang2021deeplm} to ensure a good speed. We also use the trick of discreting and interpolation to speed up. Note that the interpolation ensures the differentiability.
 
The results are shown in Table \ref{tab:optimizer}. We can see that our method can converge to a good local minima much faster than other two popular gradient-based optimizers. With more iterations (and particle numbers), our method can still outperform others with a comparable speed. Therefore, we use this fast and efficient algorithm to solve all of the optimization problems in our paper, except the optimization for the latent code of DeepSDF.

\begin{table}[htbp]
    \centering
    \caption{\textbf{Comparison of Different Optimizer.}'*' means using more iterations (and particle numbers).}
    \resizebox{\columnwidth}{!}{
    \begin{tabular}{c|ccc|ccc|ccc|c}
        \toprule
        & \multicolumn{3}{c|}{bottle}  & \multicolumn{3}{c|}{box}   & \multicolumn{3}{c|}{others*} & Speed  \\
         & 5$^\circ$5cm & 10$^\circ$10cm & CD & 5$^\circ$5cm & 10$^\circ$10cm & CD & 5$^\circ$5cm & 10$^\circ$10cm & CD & (FPS)\\
         \midrule
        Adam \cite{kingma2014adam} &38.9&55.9&2.30& 35.0& 65.1& 2.07& 49.8&75.9&1.44& 10.4\\
      
        DeepLM \cite{huang2021deeplm}  & 38.1 & 59.5 & 2.28
        &48.2&78.5&\textbf{1.74}
        &34.6&69.6&1.68
        & 7.4\\
        
        Ours  & \textbf{43.7} & \textbf{68.9} & \textbf{2.06} & \textbf{54.5} & \textbf{84.6} & \textbf{1.74} & \textbf{57.0}& \textbf{82.3}& \textbf{1.20} & \textbf{29.1}\\
        \midrule
        Adam*   & \textbf{51.3} & \textbf{72.0} & 2.06& 52.8& 84.3& \textbf{1.69}&52.1 &78.9&1.63& 4.2\\
        DeepLM*  & 40.7&61.5&2.07&50.8&82.8&\textbf{1.69}&33.5&67.4&1.76&\textbf{5.9}\\
       
        Ours* & 49.0 & \textbf{72.0} & \textbf{2.04} & \textbf{56.2} & \textbf{87.1} & 1.75 & \textbf{57.9}&\textbf{83.3}& \textbf{1.21} & 5.3\\
        \bottomrule
    \end{tabular}
    }
    \label{tab:optimizer}
\end{table}

\section{Discussion about Hand Shape Update}
\label{sec: hand shape update}
\subsection{Whether We Should Update Hand Shape}
In this section, we discuss the hand shape update during tracking. In our main paper, we use the hand shape estimated at frame $0$ in the whole video due to the simplicity and the visual consistency. Actually, we can update the hand shape over time, which may relax the dependency of the observation at frame $0$ and give us a better estimation. Through experiments, we find that updating hand shape can steadily reduce the mean per joint error.

\subsection{Method for Hand Shape Update}




For every 10 frames, we update the hand shape $\boldsymbol{\beta}$ via optimizing the following energy:
\begin{equation}
    E_t(\boldsymbol{\beta}) = \frac{1}{|T|}\sum_{t'\in T} \| l(\J^{pred}_{t'}) - l(\J^{MANO}(\mathbf{0}, \boldsymbol{\beta})) \|_1
\end{equation}

where $\J^{pred}_{t'}$ is the hand joint prediction at frame $t'$ and $T=\{i|i<t, i\%10=0\}$. By optimizing this energy, we make sure the estimated hand shape code $\boldsymbol{\beta}$ is the best match of all observations till now.

\subsection{Experiment and Analysis}
We report the experiments of updating the hand shape in Table \ref{tab:hand shape} and show the performance of using ground-truth hand shape for reference. We find the hand shape update can improve our performance marginally. Comparing the last two rows,  we can also see that after adding this hand shape update technique, there is not much room left for a better hand shape to improve the final hand joint error.

\begin{table}[htbp]
    \centering
    \caption{\textbf{Update hand shape during tracking.} }
    \resizebox{0.7\columnwidth}{!}{
    \begin{tabular}{c|c|c|c}
        \toprule
        MPJPE (cm)$\downarrow$ & \multicolumn{1}{c|}{SimGrasp}  & \multicolumn{1}{c|}{HO3D}   & \multicolumn{1}{c}{DexYCB} \\
         \midrule
        w/o shape update & 0.96 & 2.34 & 2.99\\
        w/ shape update & \textbf{0.92} & \textbf{2.26} & \textbf{2.89}\\
        \midrule
        w/ ground-truth hand shape & 0.89 & 2.24 & 2.89\\
        \bottomrule
    \end{tabular}
    }
    \label{tab:hand shape}
\end{table}

\vspace{-3mm}
\section{Discussion about Object Shape Update}
\label{sec: object shape update}
\vspace{-1mm}
\subsection{Whether We Should Update Object Shape}
Again, for the same reasons of simplicity as in Sec.\ref{sec: hand shape update}, we don't mention object shape update in the main paper and attach the discussion about object shape updating here.

\vspace{-1mm}
\subsection{Method for Shape Update}

Generally speaking, we aggregate the canonicalized object point clouds of each frame, and optimize the latent code of DeepSDF to update the object shape every few frames. In the following we will explain in details. 

At each frame $t$, after estimating the object pose $\{\R_t^{obj}, \T_t^{obj}\}$ and transform the observed point cloud $_C\PC^{obj}_t$ to object coordinate space by $_O\PC^{obj}_t=(\R_t^{obj})^{-1}(_C\PC^{obj}_t-\T_t^{obj})$
, we aggregate $_O\PC^{obj}_t$ to maintain a history point cloud $_O\PC^{his}_{t}$ by forgetting old points and adding new points as follows

\begin{equation}
    _O\PC_{t}^{his}~=~S_1~\cup~S_2
\end{equation}

in which $S_1$ is a random $\frac{(t-1)N}{t}$-subset of $_O\PC_{t-1}^{his}$, and $S_2$ is a random $\frac{N}{t}$-subset of $\{\p~|~\p\in~_O\PC_{t}^{obj}~\land~-\lambda<\psi(\p) < \lambda \}$. $\psi(\p)$ is the signed distance of $\p$ to the current object surface. We avoid introducing new noise by throwing away the points which are far from the current object surface, and mitigate the impact of previous noisy points by random forgetting.

After each k frames (we set k=10), we update the object shape by optimizing the latent code of DeepSDF in an offline manner. Once the shape update is done, we can use the updated object shape for the following tracking.

\vspace{-3mm}
\subsection{Experiment and Analysis}
\vspace{-3mm}
As shown in Table \ref{table: update shape}, we can see that the shape update can often bring benefits, especially for recovering the details of shape, which in turn benefits the pose estimation task. However, sometimes the shape may become worse due to the large pose error.

Another interesting phenomena is that the shape update can bring more benefit in HO3D than in DexYCB. This is because the video in HO3D is much longer and one video often contains more different views of a object than in DexYCB. What's more, the motion blur of DexYCB challenges the pose estimation task purely from depth point clouds. We can see in Table  \ref{table: update shape} that in DexYCB even if we use the ground truth object shape, the accuracy of pose can only be improved marginally, indicating that the object shape may not be the bottleneck here.

Finally we analyze the time cost of our offline object shape update. It takes 0.5s to optimize the latent code of DeepSDF with Adam optimizer, and it takes another 0.5s to update the SDF grids, introduced in \ref{sec: particle swarm}, for the object pose optimization. Therefore, the total time cost to update the object shape is 1.0s each time. Since our tracking system runs at 9FPS, we can finish the shape update within 10 frames, therefore  update the shape no frequent than once per 10 frames would not lead to any time cost.

\begin{table}[htbp]
    \caption{\textbf{Update object shape during tracking.} Left: '$^\dagger$' denotes that the object category has symmetry. Right: Comparison of initial shape and updated shape. The column in the red rectangular is the failure case caused by pose error.}
\begin{minipage}[h]{0.72\columnwidth}
    \resizebox{\columnwidth}{!}{
        \begin{tabular}{cc|cc|cccc}
        \toprule 
        &&\multicolumn{2}{|c|}{HO3D} &
        \multicolumn{4}{c}{DexYCB} \\
        & & bottle & box  &bottle &  bowl$^\dagger$ &  box & can$^\dagger$  \\
        \midrule
        
        w/o & 5$^\circ$5cm($\%$) $\uparrow$ & 
        43.7 &54.5 
         & 38.5 & \textbf{33.3 }
        &\textbf{31.2}&\textbf{35.7}
        \\
        shape & 10$^\circ$10cm($\%$) $\uparrow$
         &68.9 & 84.6
        &64.1&58.4
         &\textbf{52.0}&\textbf{59.0}\\
        update & CD(cm) $\downarrow$ & 2.06 & \textbf{1.74} 
        & 2.46&\textbf{4.71}
        &2.82&2.32\\
        
        \midrule
    
       w/ & 5$^\circ$5cm($\%$) $\uparrow$ &\textbf{50.9}&\textbf{57.2} &\textbf{39.3} &32.0&30.9& 33.7    \\
        shape & 10$^\circ$10cm($\%$) $\uparrow$ &\textbf{76.5}&\textbf{85.9}&\textbf{64.9} &\textbf{60.1}&49.1&55.9\\
        update & CD(cm) $\downarrow$&\textbf{1.79}&1.83&\textbf{2.28}&5.31&\textbf{2.56}&\textbf{2.07}\\

        \midrule
       GT & 5$^\circ$5cm($\%$) $\uparrow$ & 
        71.1 &77.6
         & 43.9&28.8&32.1
        &40.1
        \\
        shape & 10$^\circ$10cm($\%$) $\uparrow$
        & 91.7& 95.6
        &75.8&52.4&57.2&64.0\\
         & CD(cm) $\downarrow$ &1.73 & 1.39 
        & 1.40&2.44&1.61&1.71
        \\
        \bottomrule
        \end{tabular}
    }
\end{minipage}
\begin{minipage}[h]{0.25\columnwidth}
\includegraphics[width=\columnwidth]{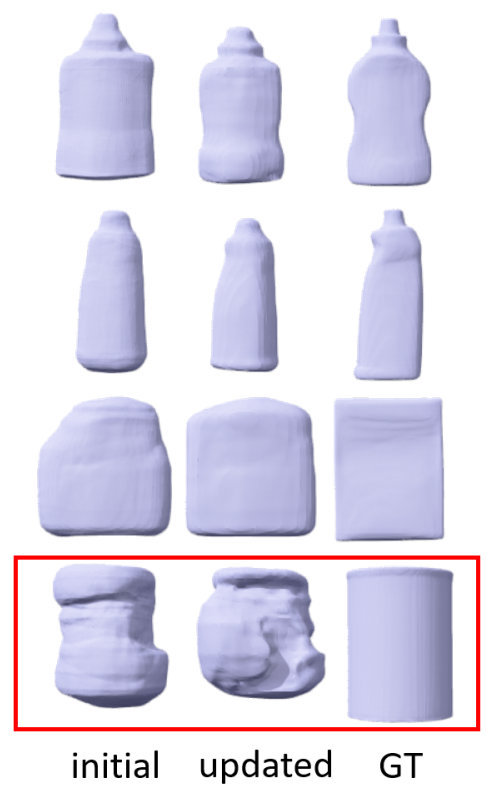}
\end{minipage}
    \label{table: update shape}
  
\end{table}

\vspace{-3mm}
\section{SimGrasp}
\label{sec: simgrasp}
\vspace{-3mm}
\subsection{Trajectory Generation}
\vspace{-2mm}
We first use GraspIt\cite{miller2004graspit} to randomly generate some grasping poses. For each grasping pose, we pull the hand away from the object and randomly choose an initial pose near the rest hand pose. Next, we interpolate between the initial pose and the grasping pose to get a trajectory, detect collision in each frame and filter out those sequences with large hand-object penetration (>5mm). We call the resulted trajectory as pre-grasping stage. After the hand has reached the grasping pose, we also add an in-grasping stage, where the hand and object don't have relative movements. The length of pre-grasping stage is set to be $k$ frames ($k$ is a random number between 40 and 60) while the length of in-grasping stage is $100-k$ frames. Along the whole trajectory, we also randomly move the camera pose to simulate the absolute movement of hand and object. As far as we know, there is no previous method who generates hand-object interaction sequences in this way.

\vspace{-3mm}
\subsection{Statistics}
\vspace{-3mm}
The objects of our SimGrasp dataset come from the ShapeNet \cite{chang2015shapenet} dataset, which contains 3 categories (\textit{i.e.} bottle, bowl, car) with 666 different objects in total. The hands are generated by randomly choosing the shape code $\boldsymbol{\beta}$ of MANO \cite{MANO:SIGGRAPHASIA:2017}, which leads to 100 different hand shapes. As a result, SimGrasp contains 666 different object instances and 100 hand shapes, which forms 1810 videos with 100 frames per video. 

\subsection{Data Visualization}
\label{sec:data_vis}
\begin{figure}[h]
    \centering
    \includegraphics[width=\columnwidth]{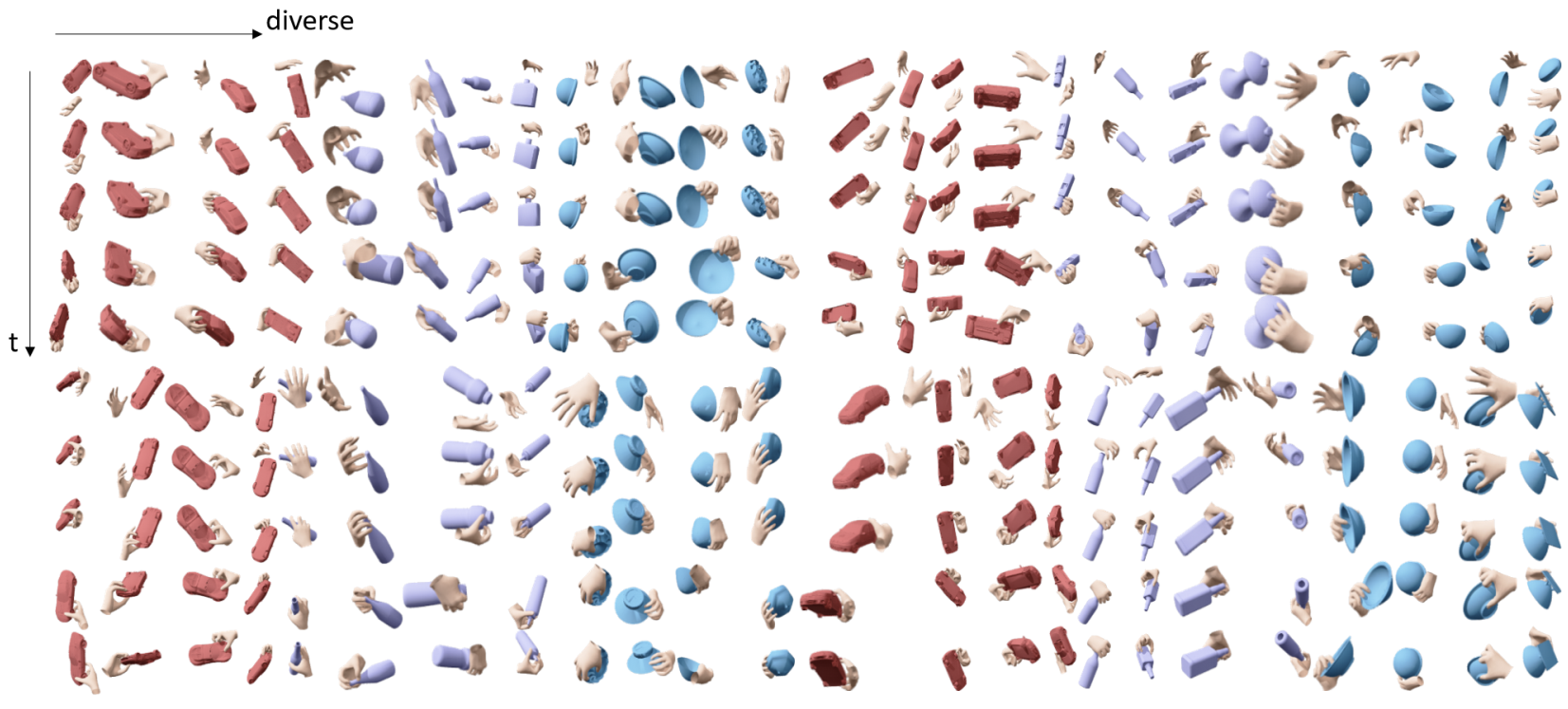}
    \caption{\textbf{Visualization for SimGrasp}. We synthesize a large-scale dynamic dataset SimGrasp with sufficient variations and realistic sensor noises.}
\end{figure}

\section{Testing Datasets}
\label{sec: testing datasets}

\begin{figure}[htbp]
    \centering
    \includegraphics[width=0.9\columnwidth]{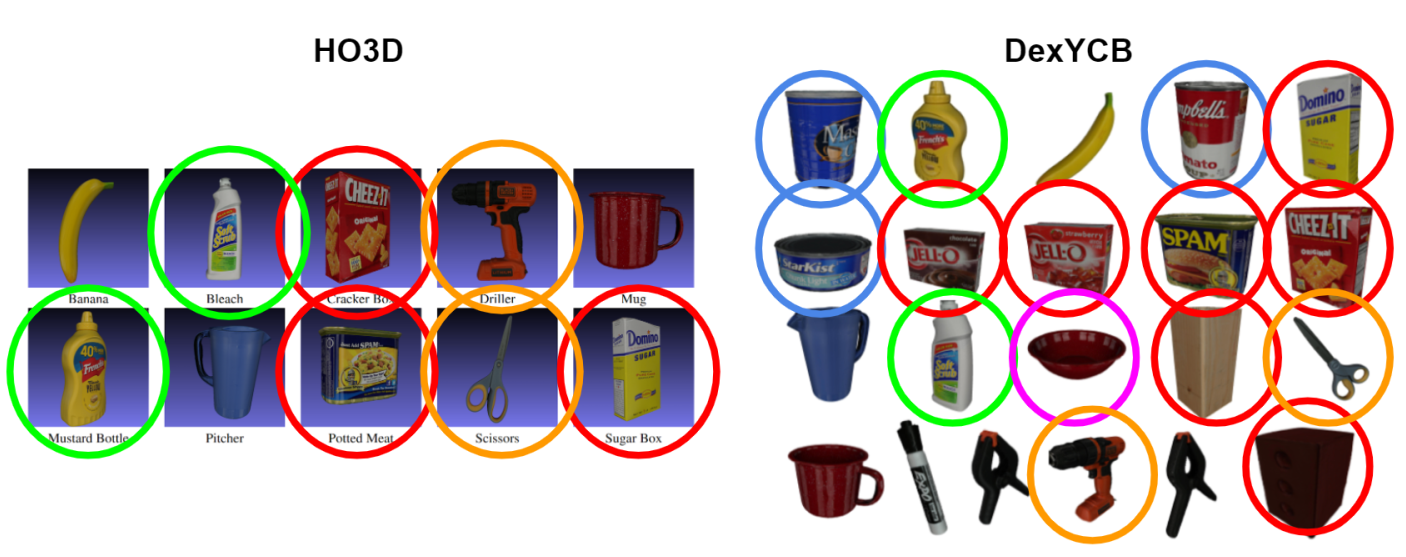}
    \caption{Category splits for HO3D and DexYCB. The instances in circles of the same color are denoted as one category. Green: 'bottle'. Red: 'box'. Blue: 'can'. Purple: 'bowl'. Orange: 'others'. }
    \label{fig:split}
\end{figure}

\textbf{HO3D}\cite{hampali2020ho3dv1} is a RGBD video dataset with accurate 3D annotation and high-quality depth images. We use the most recent version of HO3D, HO3Dv3, as our testing dataset.
Since our tracking model assumes that the initialization and the segmentation mask is known but the testing set annotation is withheld for codalab challenge, we use a subset of the official training dataset for testing and evaluation. Specifically, we manually splits the sequences in the training dataset into three categories according to the object instances, which is shown in Fig. \ref{fig:split}. Our testing set contains 118 videos and 38633 frames in total, 26 videos for bottle, 37 videos for boxes, 55 videos for others.

\textbf{DexYCB}\cite{chao2021dexycb} is another RGBD video dataset, which is very different from the popular HO3D dataset. This dataset contains much more object instances from YCB dataset, and the human subjects interact faster and more freely with objects, which greatly increases the difficulty of precise pose tracking. We follow the official $s_0$ testing split and use a subset of it for testing and evaluation, which results in 384 videos and 19868 frames in total. 
We only use a subset mainly because we force on those objects with similar geometry to the training data of our DeepSDF model. What's more, there are both left hands and right hands in DexYCB dataset, but we only focus on the right hand. The splits of each category are also shown in Fig. \ref{fig:split}.

\textbf{Data preprocess} Since our method needs point cloud sequences as input while these datasets only provide depth images, we need project them back to 3D. This back-projection can be done by the following equations: 
$$
[x\ y\ z]^\top=z\mathbf{K}^{-1}[u\ v\ 1]^\top
$$
in which $\mathbf{K}$ is the intrinsics of the depth sensor provided by the datasets, $[u\ v]$ is the 2D coordinate and $z$ is the corresponding depth. 

\vspace{-3mm}

\section{More Experiments}
\label{sec: more ablation}

\vspace{-3mm}
\subsection{More Cross-Dataset Evaluation}
\vspace{-3mm}
In our paper, we trained purely on our simulated dataset and tested on two real-world datasets HO3D and DexYCB, to show the generalization ability of our method. To demonstrate the superiority of our system, we conduct three sets of new experiments: 1) train all the methods on HO3D and test on DexYCB; 2) train all the methods on DexYCB and test on HO3D; 3) train and test all the method on DexYCB. Note that we don't train and test on HO3D because the testing set is withheld for the CodaLab challenge but we need annotations for initialization and segmentation. Therefore, we use the standard training dataset of HO3D to test the cross-dataset performance.

As shown in Table \ref{tab:cross-dataset}, no matter to train on which dataset and test on which dataset, our method is always the best and significantly outperforms the others. This validates our claim that our method has a powerful generalization capability and is robust to the choice of train data.

We also find that different methods may prefer different training data on the different testing datasets. There is no conclusion on which dataset as training data is the best. In most of the cases, our simulated dataset, SimGrasp, is comparable (in 7 over 8 experiments) or better (in 5 over 8 experiments) than real datasets. Also, the best performing model on HO3D is trained on our SimGrasp as well as the second best of DexYCB in the cross-dataset setting. Considering our SimGrasp is annotation-free and the possible future works based on this dataset, we still think it is an important contribution to this field.

\begin{table}[htbp]
    \centering
    \resizebox{\columnwidth}{!}{
    \begin{tabular}{c|ccccc}
        MPJPE(cm) & SimGrasp $\rightarrow$ DexYCB &	SimGrasp $\rightarrow$ HO3D &	HO3D $\rightarrow$ DexYCB(new) &	DexYCB $\rightarrow$ HO3D(new) &	DexYCB $\rightarrow$ DexYCB(new) \\
        \midrule
        HandFoldingNet  & 	3.78  & 	2.93  &	3.83  & 	2.65  & 	2.53 \\
        A2J  &	3.52  &	4.03  &	3.95 	 &2.53  &	1.36 \\
        VirtualView  &	3.05  &	2.71  &	3.18  &	2.83  &	2.07\\
        \midrule
        Ours  &	\textbf{2.69}  &	\textbf{2.06} &	\textbf{2.24} &	\textbf{2.14}  &	\textbf{1.21}
    \end{tabular}
    }
     \caption{ Cross-Dataset Evaluation. `A $\rightarrow$ B' means training on dataset A and testing on dataset B.}
     \label{tab:cross-dataset}
\end{table}

\vspace{-4mm}
\subsection{More Evaluation Metrics}
\vspace{-2mm}
In Table 1 of our main paper, we only report MPJPE (mean per joint position error) to evaluate the accuracy of the hand tracking. However, for tracking models, this metric may be dominated by scenarios when complete track losing happens since joint errors in that scenarios are often over 5 cm, way more larger than the common 2 cm errors. 

\begin{figure}[htbp]
    \centering
    \includegraphics[width=0.9\columnwidth]{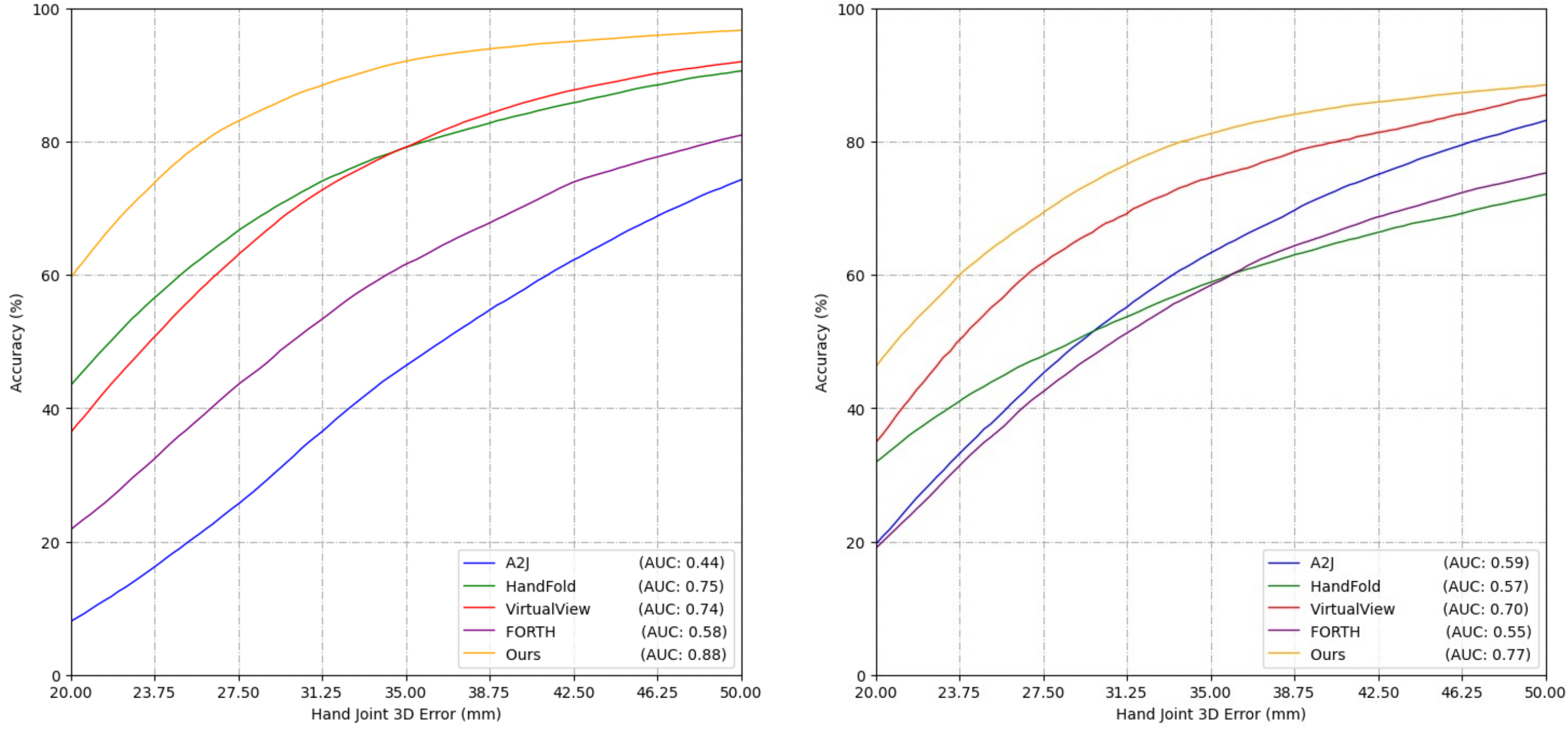}
    \caption{\textbf{Percentage of correct 3D key points (PCK) with thresholds ranging from 20mm to 50mm.} Left: PCK on HO3D. Right: PCK on DexYCB.}
    \label{fig:PCK}
\end{figure}

Therefore, to disentangle "complete track losing error" versus "on-track error", we add two more metric PCK and AUC here, following previous work \cite{kulon2020weakly, liu2021semi}. For Table 1 in our main paper, we first add the percentage of correct 3D key points (PCK) with thresholds ranging from 20mm to 50mm and draw the accuracy curve of all the methods in Fig. \ref{fig:PCK}. We then add the Area Under the Curve (AUC) metric for this curve to evaluate the mean accuracy within this error range and report the results in Table \ref{table:AUC}. Note that, these two metrics only evaluate the accuracy under 50mm, so they will not be affected by those "complete tracking losing" cases where the joint errors are usually larger than 50mm.

As shown in Table \ref{table:AUC}, our method is the best among all the methods on both datasets. Also, the curves in Fig. \ref{fig:PCK} show that our method is the best under any error threshold.

\begin{table}[htbp]
    \centering
    \caption{\textbf{AUC metric for hand pose tracking.} We show the Area Under the Curve (AUC) for the percentage of correct 3D key points (PCK) with thresholds ranging from 20mm to 50mm.}
    \resizebox{0.8\columnwidth}{!}{
        \begin{tabular}{c||ccccc}
        \toprule 
        AUC(20-50mm)$\uparrow$&\multicolumn{1}{c}{A2J\cite{xiong2019a2j}} & \multicolumn{1}{c}{HandFoldingNet\cite{cheng2021handfoldingnet}} & \multicolumn{1}{c}{VirtualView\cite{virtulview}} & \multicolumn{1}{c}{FORTH\cite{oikonomidis2011efficient}} & \multicolumn{1}{c}{Ours}\\
        \midrule
        HO3D & 0.44 & 0.75 & 0.74 & 0.58 & \textbf{0.88}\\
        DexYCB & 0.59 & 0.57 & 0.70 & 0.55 & \textbf{0.77}\\
        \bottomrule
        \end{tabular}
    }
    \label{table:AUC}
\end{table}

\vspace{-4mm}
\subsection{More Ablation Studies for Hand Tracking}

\vspace{-1mm}
In our main paper, we simply set N=2 for the trade-off between speed and performance. If we repeat them only once, the feature processing and communication will not be enough. If we repeat them three times, the joint error can be lower but the speed will be slower and the gain is marginal. 

We also verify the canonicalization module in IKNet though experiments.

\begin{table}[htbp]
\caption{Left: Ablation study about the repeating times N of KNN query and rearrangement layer in HandTrackNet. Right: Ablation study about the canonicalization in IKNet.}
\label{table:more abla}
\begin{minipage}[h]{0.5\columnwidth}
    \resizebox{\columnwidth}{!}{
        \begin{tabular}{l|cccc}
        \toprule
         &\multicolumn{3}{c}{MPJPE(cm) $\downarrow$} & Speed(FPS)\\
         & SimGrasp & HO3D & DexYCB & \\
        \midrule
        N=1 & 0.99& 2.30 & 2.82 & \textbf{30}\\
        N=3 & \textbf{0.84} & \textbf{2.03} & \textbf{2.62} & 23\\
        \midrule
        N=2 (Ours)&\textbf{0.84}&2.11&2.75&26\\
        \bottomrule
        \end{tabular}
    }
\end{minipage}
\begin{minipage}[h]{0.5\columnwidth}
    \resizebox{\columnwidth}{!}{
        \begin{tabular}{l|ccc}
        \toprule
        &\multicolumn{3}{c}{MPJPE(cm) $\downarrow$}\\
         & SimGrasp & HO3D & DexYCB  \\
        \midrule
        w/o canon. & 1.05&2.63&3.80\\
        Ours&\textbf{0.96}&\textbf{2.34}&\textbf{2.99}\\
        \bottomrule
        \end{tabular}
    }
\end{minipage}
\end{table}

\vspace{-3mm}
\subsection{Experiment Settings of Object Tracking}
\label{sec: captra setting}
Since the testing category maybe different from the training category, we use the models trained on the most similar category for evaluation. Specifically, we train our DeepSDF model on bottles from our SimGrasp dataset (Note that the object of SimGrasp comes from ShapeNet\cite{chang2015shapenet}) and test our method on bottles, boxes and cans in HO3D and DexYCB. 

However, CAPTRA will fail dramatically under the same setting because a large portion of bottles from the ShapeNet dataset is symmetric and the great ambiguity will prevent the learning-based methods from learning a reasonable strategy for the pose estimation of asymetric objects in the real world testing dataset. 

The results are shown in Table \ref{table:captra}. We can see that the CAPTRA trained on bottles can be easily get confused about the rotation. And our method outperforms both of the two settings of CAPTRA. In our main paper, we only show the results of CAPTRA trained on toy cars since it is better than CAPTRA trained on bottles.

\begin{table}[htbp]
    \caption{\textbf{Different training strategies for CAPTRA.} }
    \resizebox{\columnwidth}{!}{
        \begin{tabular}{c|ccc|ccc|ccc|ccc}
        \toprule 
        &\multicolumn{6}{|c|}{HO3D} &
        \multicolumn{6}{c}{DexYCB} \\
        \midrule
        & \multicolumn{3}{|c|}{bottle}  & \multicolumn{3}{|c|}{box}   &\multicolumn{3}{|c|}{bottle}  & \multicolumn{3}{|c}{box}   \\
       
        & 5$^\circ$5cm & 10$^\circ$10cm & CD & 5$^\circ$5cm & 10$^\circ$10cm & CD & 5$^\circ$5cm & 10$^\circ$10cm & CD & 5$^\circ$5cm & 10$^\circ$10cm & CD \\
        \midrule
       CAPTRA[car] & 15.7  &48.5 &2.84 & \textbf{67.3} &\textbf{91.9} &2.22 & 16.9& 42.3 & 3.97& 23.6 &46.5 &3.83 \\
        CAPTRA[bottle] &13.8&35.2 &2.50&10.6&20.4&2.61& 25.3&46.6& 3.40&21.9 &46.0&3.51\\
        \midrule
        Ours & \textbf{43.7}& \textbf{68.9} &\textbf{ 2.06}&54.5&84.6&\textbf{1.74}
        & \textbf{38.5}&\textbf{64.1}& \textbf{2.46}&\textbf{31.2} &\textbf{52.0}&\textbf{2.82}\\
     
        \bottomrule
        \end{tabular}
    }
    \label{table:captra}
\end{table}

\subsection{Object Pose Update via H-O Optimization}

We have tried to jointly optimize hand and object by adding the same energy as object pose estimation in Sec. 3.4, but the results is worse than only optimize the hand as is discussed in main paper. The main reason is that the object has less freedom and it is hard for hand to provide more information about where the object should be. In addition, more variables to optimize will enlarge the search space, increasing the difficulty of optimization and slowing down the pipeline. Therefore, we freeze the object pose and only optimize the hand pose in our main paper.

\begin{table}[htbp]
    \centering
    \caption{\textbf{Update object pose in h-o optimization.} We show the results on bottle in HO3D. }
    \resizebox{0.8\columnwidth}{!}{
        \begin{tabular}{c|ccc|ccc}
        \toprule 
        &\multicolumn{3}{|c|}{Object} &
        \multicolumn{3}{c}{Hand} \\

        & 5$^\circ$5cm & 10$^\circ$10cm & CD(cm) &MPJPE(cm)&PD(cm)&DD(cm) \\
        \midrule
        w/o h-o opt. & \textbf{43.7}& \textbf{68.9} & \textbf{ 2.06} &2.14&1.86&0.97\\
        jointly h-o opt. & 40.1 & 65.2 & 2.21&1.95&1.37&0.78\\
        only opt. hand & \textbf{43.7}& \textbf{68.9} & \textbf{ 2.06}&\textbf{1.85}&\textbf{1.22}&\textbf{0.69}\\
        \bottomrule
        \end{tabular}
    }
    \label{table:ho opt}
   
\end{table}

\section{Results Visualization}
\label{sec:result_vis}
\begin{figure}[h]
    \centering
    \includegraphics[width=0.97\columnwidth]{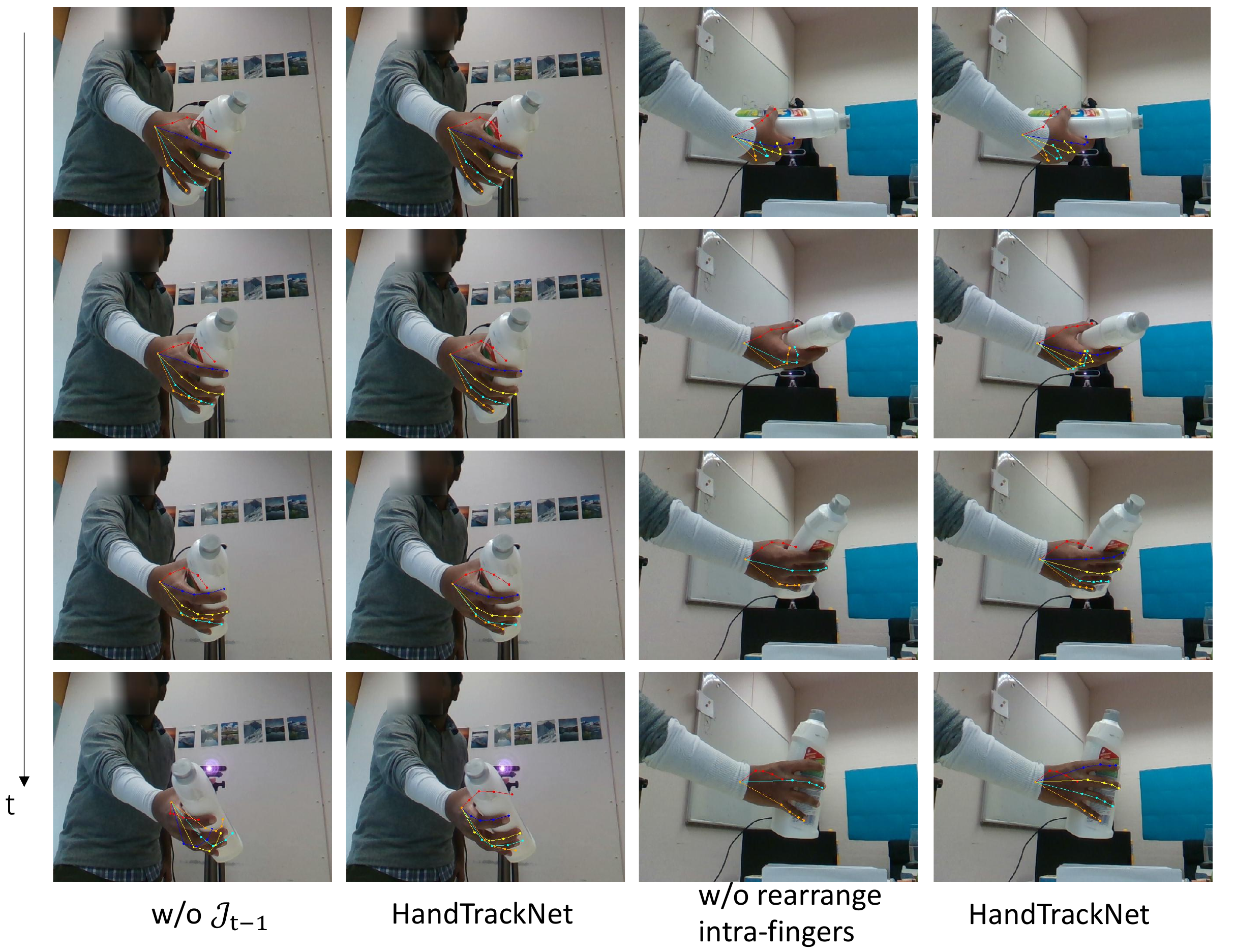}
    \caption{\textbf{Visualization for ablation study}. Left two columns: Without $\J_{t-1}$ the network will be confused under self-occlusion, \textit{e.g.} the little finger here. Right two columns: Without rearrangement among different fingers, the network will mistakenly put two fingers together.}
\end{figure}
\vspace{-6mm}

\begin{figure}[htbp]
    \centering
    \includegraphics[width=\columnwidth]{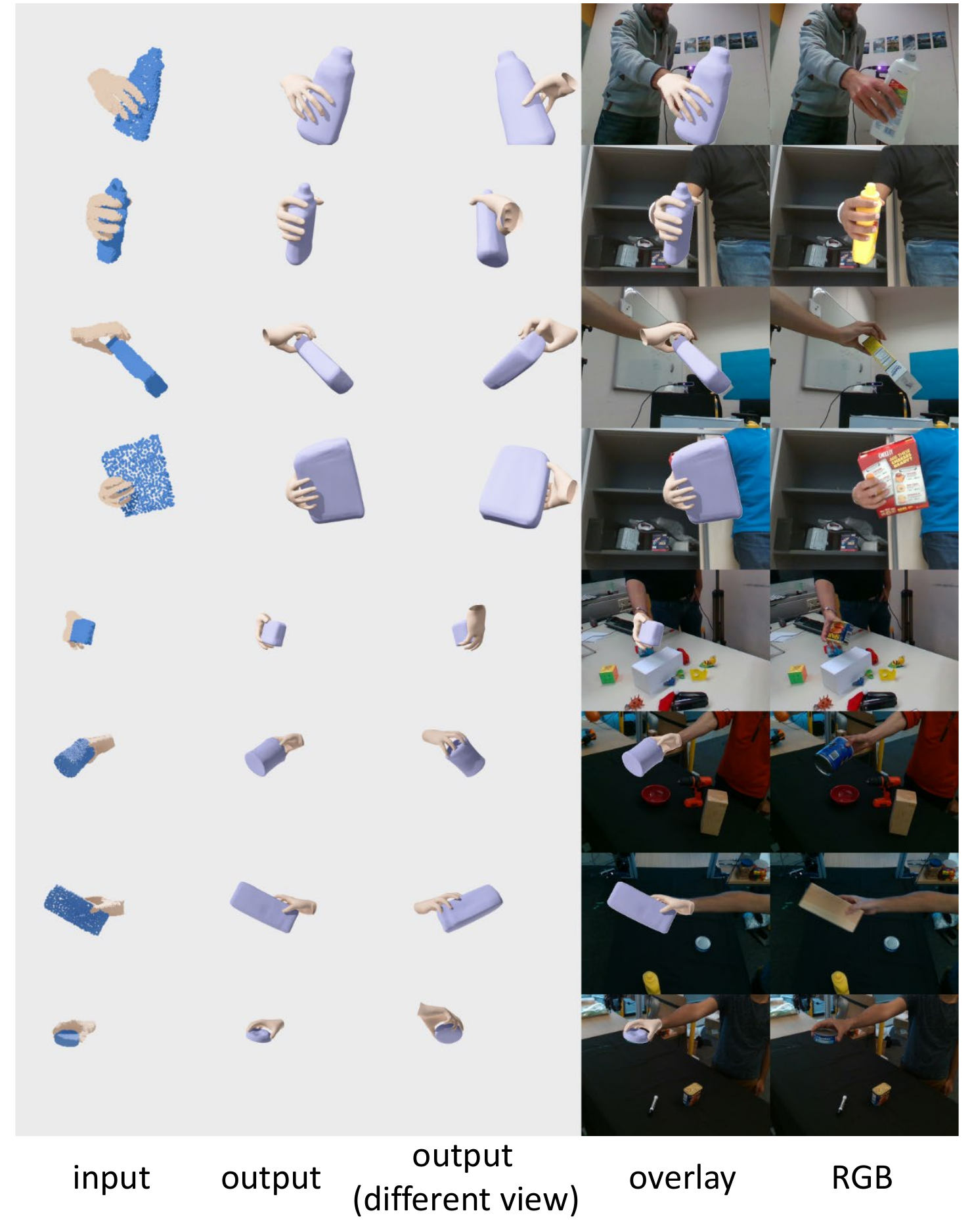}
    \caption{\textbf{Visualization on HO3D and DexYCB dataset}.}
\end{figure}

\newpage 
\section{Limitation and Social Impact}
\label{sec: limit}
Finally, we discuss about the limitation and social impact of our paper here. The robustness of the proposed model under low-quality scenarios like fast motion is one of the limitation of our system. Another limitation is that we can't reconstruct the object from a very different category (\textit{i.e.} with very different geometry) to our training data. 

The potential societal impact could be the low accuracy-related issues caused by the model performance in real applications especially those that include real humans. The carbon emission due to the network training could be another negative impact on the environment. 

\end{document}